\documentclass[journal]{IEEEtran}

\usepackage{cite}
\usepackage[pdftex]{graphicx}
\usepackage{subfigure}
\DeclareGraphicsExtensions{.pdf,.jpeg,.png,.eps,.jpg}
\usepackage{amsmath}
\usepackage{multirow}
\usepackage{array}
\usepackage{url}
\usepackage{color}
\usepackage{mathtools}
\usepackage{booktabs}
\usepackage{multirow}
\usepackage{algpseudocode}
\usepackage{algorithmicx}
\usepackage[ruled]{algorithm2e} 

\newcommand{\transpose}{\mbox{${}^{\text{T}}$}}

\newcommand\inv[1]{#1\raisebox{1.15ex}{$\scriptscriptstyle-\!1$}}

\begin{document}
\title{VINS-Mono: A Robust and Versatile Monocular Visual-Inertial State Estimator}

\author{Tong Qin, Peiliang Li, and Shaojie Shen
\thanks{T. Qin, P. Li, and S. Shen are with the Department of Electronic and Computer Engineering, Hong Kong University of Science and Technology.
        e-mail: \{tqinab, pliap\}@connect.ust.hk, eeshaojie@ust.hk}}
\maketitle

\begin{abstract}
    A monocular visual-inertial system (VINS), consisting of a camera and a low-cost inertial measurement unit (IMU), forms the minimum sensor suite for metric six degrees-of-freedom (DOF) state estimation.
    However, the lack of direct distance measurement poses significant challenges in terms of IMU processing, estimator initialization, extrinsic calibration, and nonlinear optimization.
    In this work, we present VINS-Mono: a robust and versatile monocular visual-inertial state estimator.
    Our approach starts with a robust procedure for estimator initialization and failure recovery.
    A tightly-coupled, nonlinear optimization-based method is used to obtain high accuracy visual-inertial odometry by fusing pre-integrated IMU measurements and feature observations.
    A loop detection module, in combination with our tightly-coupled formulation, enables relocalization with minimum computation overhead.
    We additionally perform four degrees-of-freedom pose graph optimization to enforce global consistency.
    We validate the performance of our system on public datasets and real-world experiments and compare against other state-of-the-art algorithms.
    We also perform onboard closed-loop autonomous flight on the MAV platform and port the algorithm to an iOS-based demonstration. 
    We highlight that the proposed work is a reliable, complete, and versatile system that is applicable for different applications that require high accuracy localization.
    We open source our implementations for both PCs\footnote{https://github.com/HKUST-Aerial-Robotics/VINS-Mono} and iOS mobile devices\footnote{https://github.com/HKUST-Aerial-Robotics/VINS-Mobile}. 
\end{abstract}

\begin{IEEEkeywords}
    Monocular visual-inertial systems, state estimation, sensor fusion, simultaneous localization and mapping 
\end{IEEEkeywords}

\section{Introduction}

\begin{figure}
    \centering
    \subfigure[Trajectory (blue) and feature locations (red)]{
        \label{fig:vins_trajectory0}    
        \includegraphics[width=0.9\columnwidth]{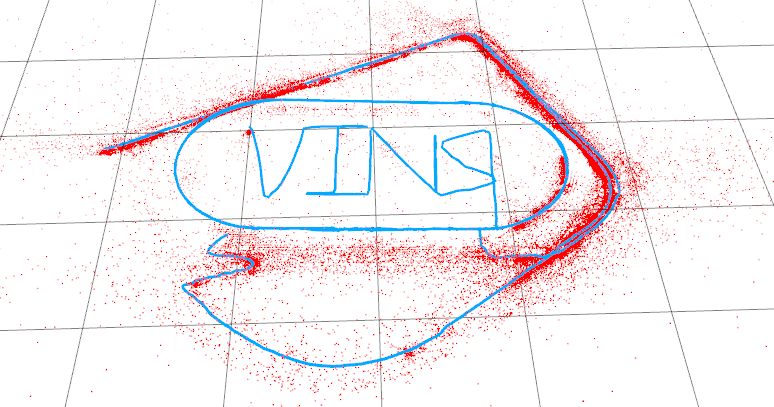}}   
    \subfigure[Trajectory overlaid with Google Map for visual comparison]{
        \label{fig:vins_map}
        \includegraphics[width=0.9\columnwidth]{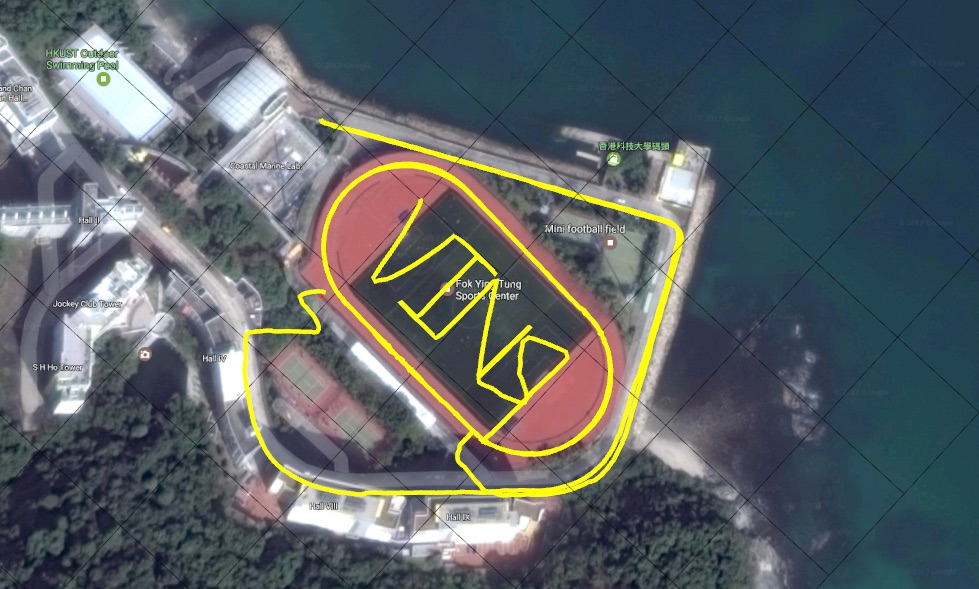}}
    \caption{Outdoor experimental results of the proposed monocular visual-inertial state estimator. 
             Data is collected by a hand-held monocular camera-IMU setup under normal walking condition.
             It includes two complete circles inside the field and two semicircles on the nearby driveway. 
             Total trajectory length is 2.5 km.
             A video of the experiment can be found in the multimedia attachment.}
    \label{fig:abstract_figure}
\end{figure}

\IEEEPARstart
{S}{tate} estimation is undoubtedly the most fundamental module for a wide range of applications, such as robotic navigation, autonomous driving, virtual reality (VR), and augmented reality (AR).
Approaches that use only a monocular camera have gained significant interests by the community due to their small size, low-cost, and easy hardware setup~\cite{klein2007parallel, ForPizSca1405, engel2014lsd, mur2015orb, engel2017direct}.
However, monocular vision-only systems are incapable of recovering the metric scale, therefore limiting their usage in real-world robotic applications.
Recently, we see a growing trend of assisting the monocular vision system with a low-cost inertial measurement unit (IMU).
The primary advantage of this monocular visual-inertial system (VINS) is to have the metric scale, as well as roll and pitch angles, all observable.
This enables navigation tasks that require metric state estimates.
In addition, the integration of IMU measurements can dramatically improve motion tracking performance by bridging the gap between losses of visual tracks due to illumination change, texture-less area, or motion blur.
In fact, the monocular VINS not only widely available on mobile robots, drones, and mobile devices, it is also the minimum sensor setup for sufficient self and environmental perception. 

However, all these advantages come with a price. For monocular VINS, it is well known that acceleration excitation is needed to make metric scale observable.
This implies that monocular VINS estimators cannot start from a stationary condition, but rather launch from an unknown moving state.
Also recognizing the fact that visual-inertial systems are highly nonlinear, we see significant challenges in terms of estimator initialization.
The existence of two sensors also makes camera-IMU extrinsic calibration critical. 
Finally, in order to eliminate long-term drift within an acceptable processing window, a complete system that includes visual-inertial odometry, loop detection, relocalization, and global optimization has to be developed. 

To address all these issues, we propose VINS-Mono, a robust and versatile monocular visual-inertial state estimator.
Our solution starts with on-the-fly estimator initialization. The same initialization module is also used for failure recovery.
The core of our solution is a robust monocular visual-inertial odometry (VIO) based on tightly-coupled sliding window nonlinear optimization.
The monocular VIO module not only provides accurate local pose, velocity, and orientation estimates, it also performs camera-IMU extrinsic calibration and IMU biases correction in an online fashion.
Loops are detected using DBoW2~\cite{GalvezTRO12}.
Relocalization is done in a tightly-coupled setting by feature-level fusion with the monocular VIO. 
This enables robust and accurate relocalization with minimum computation overhead.
Finally, geometrically verified loops are added into a pose graph, and thanks to the observable roll and pitch angles from the monocular VIO, a four degrees-of-freedom (DOF) pose graph is performed to ensure global consistency.

VINS-Mono combines and improves the our previous works on monocular visual-inertial fusion~\cite{SheMicKum1505,yang2017monocular,QinShen17,LiShen17}.
It is built on top our tightly-coupled, optimization-based formulation for monocular VIO~\cite{SheMicKum1505,yang2017monocular}, and incorporates the improved initialization procedure introduced in~\cite{QinShen17}.
The first attempt of porting to mobile devices was given in~\cite{LiShen17}.
Further improvements of VINS-Mono comparing to our previous works include 
improved IMU pre-integration with bias correction, tightly-coupled relocalization, global pose graph optimization, extensive experimental evaluation, and a robust and versatile open source implementation.

The whole system is complete and easy-to-use. It has been successfully applied to small-scale AR scenarios, medium-scale drone navigation, and large-scale state estimation tasks.
Superior performance has been shown against other state-of-the-art methods.
To this end, we summarize our contributions as follow:
\begin{itemize}
    \item A robust initialization procedure that is able to bootstrap the system from unknown initial states.
    \item A tightly-coupled, optimization-based monocular visual-inertial odometry with camera-IMU extrinsic calibration and IMU bias estimation.
    \item Online loop detection and tightly-coupled relocalization.
    \item Four DOF global pose graph optimization.
    \item Real-time performance demonstration for drone navigation, large-scale localization, and mobile AR applications.
    \item Open-source release for both the PC version that is fully integrated with ROS, as well as the iOS version that runs on iPhone6s or above.
\end{itemize}

The rest of the paper is structured as follows. 
In Sect.~\ref{sec:Related work}, we discuss relevant literature. 
We give an overview of the complete system pipeline in Sect.~\ref{sec:Overview}. 
Preprocessing steps for both visual and pre-integrated IMU measurements are presented in Sect.~\ref{sec:Measurement preprocessing}.
In Sect.~\ref{sec:Initialization}, we discuss the estimator initialization procedure.
A tightly-coupled, self-calibrating, nonlinear optimization-based monocular VIO, is presented in Sect.~\ref{sec:Optimization}.
Tightly-coupled relocalization and global pose graph optimization are presented in  Sect.~\ref{sec:rolocalization} and Sect.~\ref{sec:pose graph optimization} respectively.
Implementation details and experimental results are shown in Sect.~\ref{sec:experiment}.
Finally, the paper is concluded with a discussion and possible future research directions in Sect.~\ref{sec:conclusion}.

\begin{figure*}
    \centering
    \includegraphics[width=0.99\textwidth]{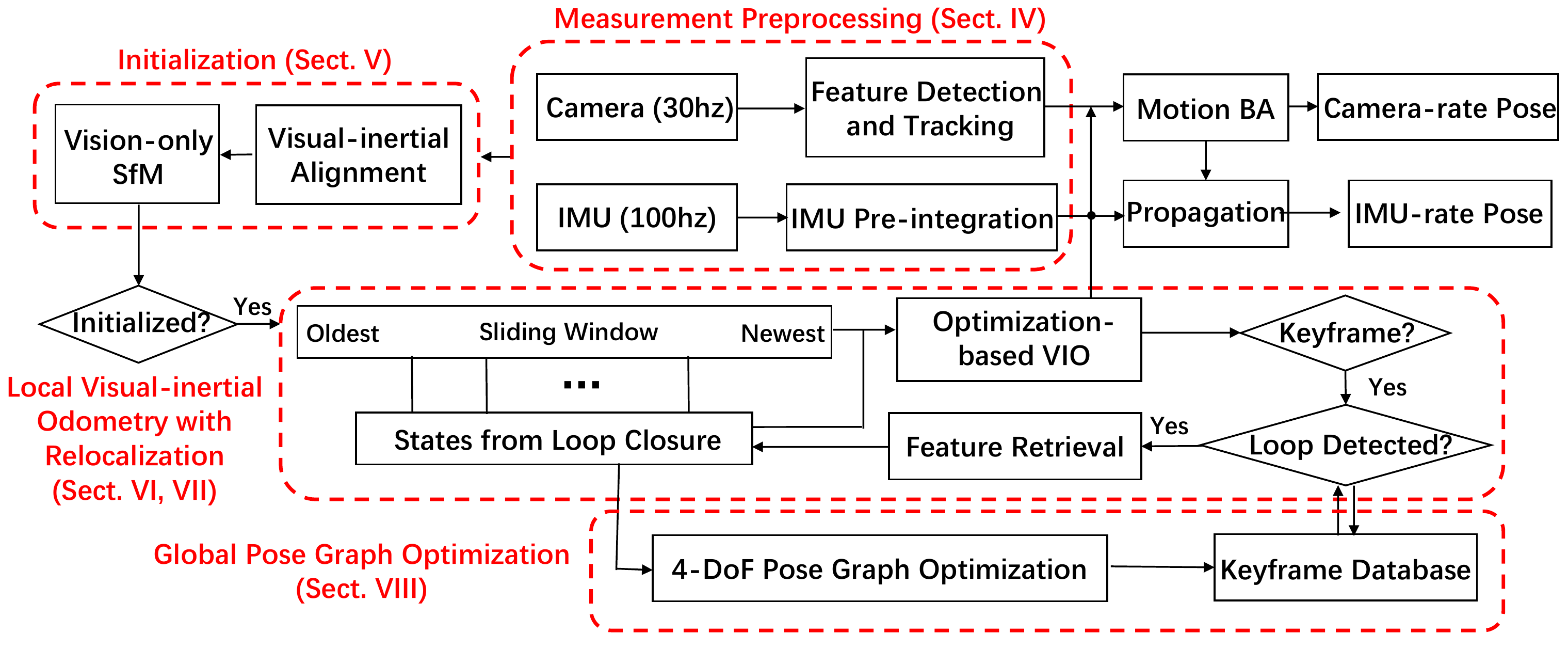}
    \caption{A block diagram illustrating the full pipeline of the proposed monocular visual-inertial state estimator.
        \label{fig:system}}
\end{figure*}

\section{Related Work}
\label{sec:Related work}
Scholarly works on monocular vision-based state estimation/odometry/SLAM are extensive.
Noticeable approaches include PTAM~\cite{klein2007parallel}, SVO~\cite{ForPizSca1405}, LSD-SLAM~\cite{engel2014lsd}, DSO~\cite{engel2017direct}, and ORB-SLAM~\cite{mur2015orb}.
It is obvious that any attempts to give a full relevant review would be incomplete.
In this section, however, we skip the discussion on vision-only approaches, and only focus on the most relevant results on monocular visual-inertial state estimation.

The simplest way to deal with visual and inertial measurements is loosely-coupled sensor fusion~\cite{weiss2012real,lynen2013robust}, where IMU is treated as an independent module to assist vision-only pose estimates obtained from the visual structure from motion. 
Fusion is usually done by an extended Kalman filter (EKF), where IMU is used for state propagation and the vision-only pose is used for the update.
Further on, tightly-coupled visual-inertial algorithms are either based on the EKF~\cite{MouRou0704,LiMou1305,bloesch2015robust} or graph optimization~\cite{LeuFurRab1306,SheMicKum1505,yang2017monocular,mur2016visual}, 
where camera and IMU measurements are jointly optimized from the raw measurement level.
A popular EKF based VIO approach is MSCKF~\cite{MouRou0704,LiMou1305}. MSCKF maintains several previous camera poses in the state vector, and uses visual measurements of the same feature across multiple camera views to form multi-constraint update.
SR-ISWF \cite{wu2015square,paulcomparative} is an extension of MSCKF. It uses square-root form \cite{kaess2012isam2} to achieve single-precision representation and avoid poor numerical properties. 
This approach employs the inverse filter for iterative re-linearization, making it equal to optimization-based algorithms.
Batch graph optimization or bundle adjustment techniques maintain and optimize all measurements to obtain the optimal state estimates.
To achieve constant processing time, popular graph-based VIO methods~\cite{LeuFurRab1306,yang2017monocular,mur2016visual} usually optimize over a bounded-size sliding window of recent states by marginalizing out past states and measurements. 
Due to high computational demands of iterative solving of nonlinear systems, few graph-based can achieve real-time performance on resource-constrained platforms, such as mobile phones.

For visual measurement processing, algorithms can be categorized into either direct or indirect method according to the definition of visual residual models.
Direct approaches \cite{ForPizSca1405,engel2014lsd,usenko2016direct} minimize photometric error while indirect approaches \cite{LeuFurRab1306,yang2017monocular,LiMou1305} minimize geometric displacement.
Direct methods require a good initial guess due to their small region of attraction, while indirect approaches consume extra computational resources on extracting and matching features.
Indirect approaches are more frequently found in real-world engineering deployment due to its maturity and robustness.
However, direct approaches are easier to be extended for dense mapping as they are operated directly on the pixel level.

In practice, IMUs usually acquire data at a much higher rate than the camera.
Different methods have been proposed to handle the high rate IMU measurements.
The most straightforward approach is to use the IMU for state propagation in EKF-based approaches ~\cite{weiss2012real,MouRou0704}.
In a graph optimization formulation, an efficient technique called IMU pre-integration is developed in order to avoid repeated IMU re-integration
This technique was first introduced in~\cite{LupSuk1202}, which parametrize rotation error using Euler angles. 
An on-manifold rotation formulation for IMU-preintegration was developed in our previous work~\cite{SheMicKum1505}.
This work derived the covariance propagation using continuous-time IMU error state dynamics. 
However, IMU biases were ignored. 
The pre-integration theory was further improved in~\cite{ForCarDel1507} by adding posterior IMU bias correction. 

Accurate initial values are crucial to bootstrap any monocular VINS.
A linear estimator initialization method that leverages relative rotations from short-term IMU pre-integration was proposed in~\cite{SheMulMic1406,yang2017monocular}.
However, this method does not model gyroscope bias, and fails to model sensor noise in raw projection equations. 
In real-world applications, this results in unreliable initialization when visual features are far away from the sensor suite.
A closed-form solution to the monocular visual-inertial initialization problem was introduced in~\cite{Mar1308}. 
Later, an extension to this closed-form solution by adding gyroscope bias calibration was proposed in~\cite{kaiser2017simultaneous}. 
These approaches fail to model the uncertainty in inertial integration since they rely on the double integration of IMU measurements over an extended period of time.
In~\cite{faessler2015automatic}, a re-initialization and failure recovery algorithm based on SVO~\cite{ForPizSca1405} was proposed. 
This is a practical method based on a loosely-coupled fusion framework.
However, an additional downward-facing distance sensor is required to recover the metric scale.
An initialization algorithm built on top of the popular ORB-SLAM~\cite{mur2015orb} was introduced in~\cite{mur2016visual}.
An initial estimation of the scale, gravity direction, velocity and IMU biases are computed for the visual-inertial full BA given a set of keyframes from ORB-SLAM.
However, it is reported that the time required for scale convergence can be longer than 10 seconds. 
This can pose problems for robotic navigation tasks that require scale estimates right at the beginning.

VIO approaches, regardless the underlying mathematical formulation that they rely on, suffer from long term drifting in global translation and orientation.
To this end, loop closure plays an important role for long-term operations.
ORB-SLAM~\cite{mur2015orb} is able to close loops and reuse the map, which takes advantage of Bag-of-World~\cite{GalvezTRO12}. 
A 7 DOF~\cite{strasdat2010scale} (position, orientation, and scale) pose graph optimization is followed loop detection. 
In contrast, for monocular VINS, thanks to the addition of IMU, drift only occurs in 4 DOF, which is the 3D translation, and the rotation around the gravity direction (yaw angle).
Therefore, in this paper, we choose to optimize the pose graph with loop constraints in the minimum 4 DOF setting.

\section{Overview}
\label{sec:Overview}
The structure of proposed monocular visual-inertial state estimator is shown in Fig.~\ref{fig:system}. 
The system starts with measurement preprocessing (Sect.~\ref{sec:Measurement preprocessing}), in which features are extracted and tracked, and IMU measurements between two consecutive frames are pre-integrated.
The initialization procedure (Sect.~\ref{sec:Initialization} provides all necessary values, including pose, velocity, gravity vector, gyroscope bias, and 3D feature location, for bootstrapping the subsequent nonlinear optimization-based VIO.
The VIO (Sect.~\ref{sec:Optimization}) with relocalization (Sect.~\ref{sec:rolocalization}) modules tightly fuses pre-integrated IMU measurements, feature observations, and re-detected features from loop closure.
Finally, the pose graph optimization module (Sect.~\ref{sec:pose graph optimization}) takes in geometrically verified relocalization results, and perform global optimization to eliminate drift.
The VIO, relocalization, and pose graph optimization modules run concurrently in a multi-thread setting.
Each module has different running rates and real-time guarantees to ensure reliable operation at all times. 

We now define notations and frame definitions that we use throughout the paper.
We consider $(\cdot)^w$ as the world frame.
The direction of the gravity is aligned with the $z$ axis of the world frame.
$(\cdot)^b$ is the body frame, which we define it to be the same as the IMU frame.
$(\cdot)^c$ is the camera frame.
We use both rotation matrices $\mathbf{R}$ and Hamilton quaternions $\mathbf{q}$ to represent rotation.
We primarily use quaternions in state vectors, but rotation matrices are also used for convenience rotation of 3D vectors.
$\mathbf{q}^w_b, \mathbf{p}^w_b$ are rotation and translation from the body frame to the world frame. 
$b_k$ is the body frame while taking the $k^{th}$ image.
$c_k$ is the camera frame while taking the $k^{th}$ image.
$\otimes$ represents the multiplication operation between two quaternions.
$\mathbf{g}^w = [0,0,g]^T$ is the gravity vector in the world frame.
Finally, we denote $\hat{(\cdot)}$ as the noisy measurement or estimate of a certain quantity.

\begin{figure*}
    \centering
    \includegraphics[width=0.9\textwidth]{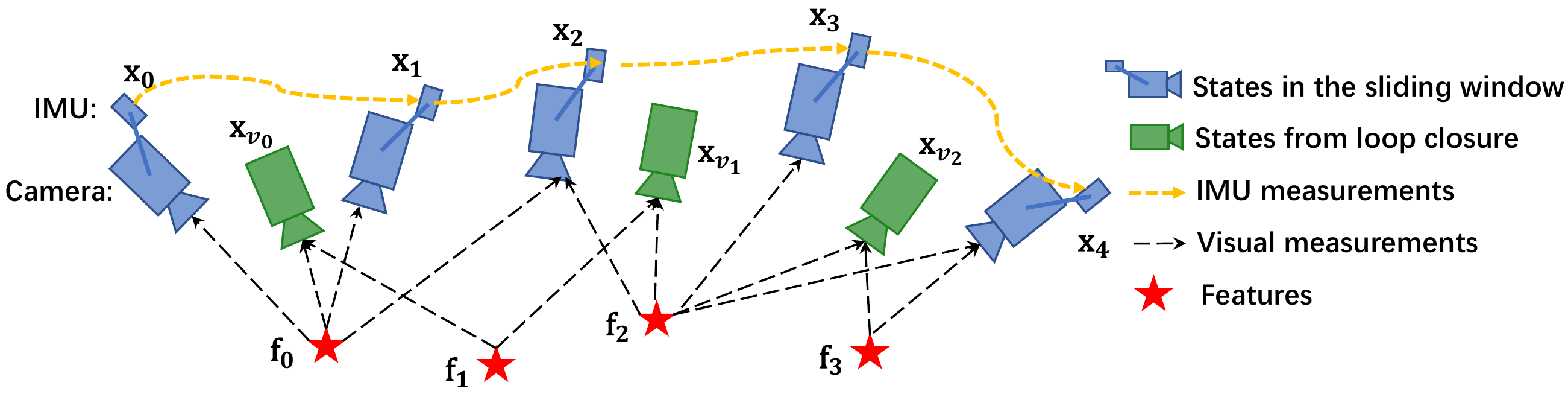}
    \caption{An illustration of the sliding window monocular VIO with relocalization. It is a tightly-coupled formulation with IMU, visual, and loop measurements.
        \label{fig:window}}
\end{figure*}

\section{Measurement Preprocessing}
\label{sec:Measurement preprocessing}
This section presents preprocessing steps for both inertial and monocular visual measurements.
For visual measurements, we track features between consecutive frames and detect new features in the latest frame.
For IMU measurements, we pre-integrate them between two consecutive frames. 
Note that the measurements of the low-cost IMU that we use are affected by both bias and noise. 
We therefore especially take bias into account in the IMU pre-integration process.

\subsection{Vision Processing Front-end}
\label{sec:vision processing}
For each new image, existing features are tracked by the KLT sparse optical flow algorithm~\cite{LucKan8108}. 
Meanwhile, new corner features are detected~\cite{shi1994good} to maintain a minimum number (100-300) of features in each image. 
The detector enforces a uniform feature distribution by setting a minimum separation of pixels between two neighboring features. 
2D Features are firstly undistorted and then projected to a unit sphere after passing outlier rejection. 
Outlier rejection is performed using RANSAC with fundamental matrix model~\cite{hartley2003multiple}.

Keyframes are also selected in this step. 
We have two criteria for keyframe selection. 
The first one is the average parallax apart from the previous keyframe. 
If the average parallax of tracked features is between the current frame and the latest keyframe is beyond a certain threshold, we treat frame as a new keyframe. 
Note that not only translation but also rotation can cause parallax.
However, features cannot be triangulated in the rotation-only motion. 
To avoid this situation, we use short-term integration of gyroscope measurements to compensate rotation when calculating parallax. 
Note that this rotation compensation is only used to keyframe selection, and is not involved in rotation calculation in the VINS formulation.
To this end, even if the gyroscope contains large noise or is biased, it will only result in suboptimal keyframe selection results, and will not directly affect the estimation quality.
Another criterion is tracking quality. If the number of tracked features goes below a certain threshold, we treat this frame as a new keyframe.
This criterion is to avoid complete loss of feature tracks.

\subsection{IMU Pre-integration}
\label{subsec:imu_pre-integration}
IMU Pre-integration was first proposed in \cite{LupSuk1202}, which parametrized rotation error in Euler angle. 
An on-manifold rotation formulation for IMU pre-integration was developed in our previous work~\cite{SheMicKum1505}.
This work derived the covariance propagation using continuous-time IMU error state dynamics. 
However, IMU biases were ignored. 
The pre-integration theory was further improved in~\cite{ForCarDel1507} by adding posterior IMU bias correction. 
In this paper, we extend the IMU pre-integration proposed in our previous work~\cite{SheMicKum1505} by incorporating IMU bias correction.  

The raw gyroscope and accelerometer measurements from IMU, $\hat{\boldsymbol{\omega}}$ and $\hat{\mathbf{a}}$, are given by:
\begin{equation}
\begin{split}
\hat{\mathbf{a}}_t &=  {\mathbf{a}}_t + \mathbf{b}_{a_t} + {\mathbf{R}^t_w} \mathbf{g}^w + \mathbf{n}_a
\\
\hat{\boldsymbol{\omega}}_t &=  {\boldsymbol{\omega}}_t + \mathbf{b}_{w_t} + \mathbf{n}_w 
.
\end{split} 
\end{equation}
IMU measurements, which are measured in the body frame, combines the force for countering gravity and the platform dynamics, and are affected by acceleration bias $\mathbf{b}_a$, gyroscope bias $\mathbf{b}_w$, and additive noise. 
We assume that the additive noise in acceleration and gyroscope measurements are Gaussian, $ \mathbf{n}_a \sim \mathcal{N}(\mathbf{0},\boldsymbol{\sigma}_a^{2})$, $ \mathbf{n}_w \sim \mathcal{N}(\mathbf{0},\boldsymbol{\sigma}_w^{2})$. 
Acceleration bias and gyroscope bias are modeled as random walk, whose derivatives are Gaussian, $ \mathbf{n}_{b_a} \sim \mathcal{N}(\mathbf{0},\boldsymbol{\sigma}_{b_a}^{2})$, $ \mathbf{n}_{b_w} \sim \mathcal{N}(\mathbf{0},\boldsymbol{\sigma}_{b_w}^{2})$:
\begin{equation}
\dot{\mathbf{b}}_{a_t} = \mathbf{n}_{b_a},\quad \dot{\mathbf{b}}_{w_t} = \mathbf{n}_{b_w} .
\end{equation}

Given two time instants that correspond to image frames $b_k$ and $b_{k+1}$, position, velocity, and orientation states can be propagated by inertial measurements during time interval $[t_k, t_{k+1}]$ in the world frame:
\begin{equation}
\begin{split}\label{eq:integration_world}
\mathbf{p}^w_{b_{k+1}} & = \mathbf{p}^w_{b_k} + \mathbf{v}^w_{b_k} {\Delta t}_k 
\\& \qquad \quad
+ \iint_{t \in [t_k,t_{k+1}]} \left ( \mathbf{R}^w_t (\hat{\mathbf{a}}_t - \mathbf{b}_{a_t} - \mathbf{n}_a)- \mathbf{g}^{w} \right ) dt^2\\
\mathbf{v}^w_{b_{k+1}} &= \mathbf{v}^w_{b_k} + \int_{t \in [t_k,t_{k+1}]} \left ( \mathbf{R}^w_t (\hat{\mathbf{a}}_t- \mathbf{b}_{a_t} - \mathbf{n}_a) - \mathbf{g}^{w} \right ) dt\\
\mathbf{q }^{w}_{b_{k+1}} &=\mathbf{q}^w_{b_k} \otimes \int_{t \in [t_k,t_{k+1}]} \frac{1}{2} \boldsymbol{\Omega}(\hat{\boldsymbol{\omega}}_t - \mathbf{b}_{w_t} - \mathbf{n}_w)\mathbf{q}^{b_k}_t dt,
\end{split} 
\end{equation}
where 
\begin{equation}
\begin{split}
\boldsymbol{\Omega}(\boldsymbol{\omega}) = 
\begin{bmatrix}
    -\lfloor {\boldsymbol{\omega}} \rfloor_{\times} & {\boldsymbol{\omega}}\\
    -{\boldsymbol{\omega}}^T & 0
\end{bmatrix} ,
\lfloor \boldsymbol{\omega} \rfloor_{\times} =
\begin{bmatrix}
0& -\omega_z & \omega_y\\
\omega_z& 0 & -\omega_x\\
-\omega_y& \omega_x & 0
\end{bmatrix} .
\end{split}
\end{equation}
${\Delta t}_k$ is the duration between the time interval $[t_k, t_{k+1}]$. 

It can be seen that the IMU state propagation requires rotation, position and velocity of frame $b_k$.
When these starting states change, we need to re-propagate IMU measurements.
Especially in the optimization-based algorithm, every time we adjust poses, we will need to re-propagate IMU measurements between them.
This propagation strategy is computationally demanding.
To avoid re-propagation, we adopt pre-integration algorithm. 

After change the reference frame from the world frame to the local frame $b_k$,
we can only pre-integrate the parts which are related to linear acceleration $\hat{\mathbf{a}}$ and angular velocity $\hat{\boldsymbol{\omega}}$ as follows:
\begin{equation}
\begin{split}\label{eq:local_integration}
\mathbf{R}^{b_k}_w\mathbf{p}^{w}_{b_{k+1}} &= \mathbf{R}^{b_k}_w (\mathbf{p}^{w}_{b_k} + \mathbf{v}^{w}_{b_k} {\Delta t}_k - \frac{1}{2} \mathbf{g}^w {\Delta t}^2_k)+ \boldsymbol{\alpha}^{b_k}_{b_{k+1}}
\\
\mathbf{R}^{b_k}_w\mathbf{v}^{w}_{b_{k+1}} &= \mathbf{R}^{b_k}_w(\mathbf{v}^{w}_{b_k}- \mathbf{g}^w {\Delta t}_k) + \boldsymbol{\beta}^{b_k}_{b_{k+1}}
\\
\mathbf{q}^{b_k}_w \otimes \mathbf{q}^{w}_{b_{k+1}} &= \boldsymbol{\gamma}^{b_k}_{b_{k+1}},
\end{split} 
\end{equation}
where,
\begin{equation}
\label{eq:integration_preintegration}
\begin{split}
\boldsymbol{\alpha}^{b_k}_{b_{k+1}} &= \iint_{t \in [t_k,t_{k+1}]} \mathbf{R}^{b_k}_{t} (\hat{\mathbf{a}}_{t} - \mathbf{b}_{a_t} - \mathbf{n}_a) dt^2 \\
\boldsymbol{\beta}^{b_k}_{b_{k+1}}  &=  \int_{t \in [t_k,t_{k+1}]} \mathbf{R}^{b_k}_{t} (\hat{\mathbf{a}}_{t} - \mathbf{b}_{a_t} - \mathbf{n}_a) dt\\ 
\boldsymbol{\gamma}^{b_k}_{b_{k+1}} &=\int_{t \in [t_k,t_{k+1}]} \frac{1}{2} \boldsymbol{\Omega}(\hat{\boldsymbol{\omega}}_t - \mathbf{b}_{w_t} - \mathbf{n}_w) \boldsymbol{\gamma}^{b_k}_{t} dt.
\end{split}
\end{equation}
It can be seen that the pre-integration terms~\eqref{eq:integration_preintegration} can be obtained solely with IMU measurements by taking $b_k$ as the reference frame. 
$\boldsymbol{\alpha}^{b_k}_{b_{k+1}},\boldsymbol{\beta}^{b_k}_{b_{k+1}},\boldsymbol{\gamma}^{b_k}_{b_{k+1}}$ are only related to IMU biases instead of other states in $b_k$ and $b_{k+1}$. 
When the estimation of bias changes, if the change is small, we adjust $\boldsymbol{\alpha}^{b_k}_{b_{k+1}},\boldsymbol{\beta}^{b_k}_{b_{k+1}},\boldsymbol{\gamma}^{b_k}_{b_{k+1}}$ by their first-order approximations with respect to the bias, 
otherwise we do re-propagation. 
This strategy saves a significant amount of computational resources for optimization-based algorithms, since we do not need to propagate IMU measurements repeatedly.  

For discrete-time implementation, different numerical integration methods such as Euler, mid-point, RK4 integration can be applied. 
Here Euler integration is chosen to demonstrate the procedure for easy understanding (we use mid-point integration in the implementation code).

At the beginning, $\boldsymbol{\alpha}^{b_k}_{b_k},\boldsymbol{\beta}^{b_k}_{b_k}$ is $\mathbf{0}$, and $\boldsymbol{\gamma}^{b_k}_{b_k}$ is identity quaternion. 
The mean of $\boldsymbol{\alpha}, \boldsymbol{\beta}, \boldsymbol{\gamma}$ in~\eqref{eq:integration_preintegration} is propagated step by step as follows. 
Note that the additive noise terms $\mathbf{n}_a$, $\mathbf{n}_w$ are unknown, and is treated as zero in the implementation.
This results in estimated values of the pre-integration terms, marked by $\hat{(\cdot)}$:
\begin{equation}
\label{eq:propogation}
\begin{split}
\hat{\boldsymbol{\alpha}}^{b_k}_{i+1} &= \hat{\boldsymbol{\alpha}}^{b_k}_i + \hat{\boldsymbol{\beta}}^{b_k}_i\delta t + \frac{1}{2} \mathbf{R}(\hat{\boldsymbol{\gamma}}^{b_k}_i)(\hat{\mathbf{a}}_i - \mathbf{b}_{a_i}) \delta t^2  \\
\hat{\boldsymbol{\beta}}^{b_k}_{i+1} &= \hat{\boldsymbol{\beta}}^{b_k}_{i} + \mathbf{R}(\hat{\boldsymbol{\gamma}}^{b_k}_i)(\hat{\mathbf{a}}_i - \mathbf{b}_{a_i})\delta t\\
\hat{\boldsymbol{\gamma}}^{b_k}_{i+1} &= \hat{\boldsymbol{\gamma}}^{b_k}_i \otimes   
\begin{bmatrix}  
1\\
\frac{1}{2} (\hat{\boldsymbol{\omega}}_i- \mathbf{b}_{w_i})\delta t
\end{bmatrix}
\end{split}.
\end{equation}
$i$ is discrete moment corresponding to a IMU measurement within $[t_k, t_{k+1}]$. 
$\delta t$ is the time interval between two IMU measurements $i$ and $i+1$.

Then we deal with the covariance propagation. 
Since the four-dimensional rotation quaternion $\boldsymbol{\gamma}^{b_k}_t$ is over-parameterized, we define its error term as a perturbation around its mean:
\begin{equation}
\boldsymbol{\gamma}^{b_k}_t \approx \boldsymbol{\hat{\gamma}}^{b_k}_t \otimes 
\begin{bmatrix} 
1\\
\frac{1}{2}\delta\boldsymbol{\theta}^{b_k}_t
\end{bmatrix},
\end{equation}
where $\delta\boldsymbol{\theta}^{b_k}_t$ is three-dimensional small perturbation.

We can derive continuous-time linearized dynamics of error terms of~\eqref{eq:integration_preintegration}:
\begin{equation}
\begin{aligned}
\begin{bmatrix}
\delta \dot{\boldsymbol{\alpha}}^{b_k}_t \\
\delta \dot{\boldsymbol{\beta}}^{b_k}_t \\
\delta \dot{\boldsymbol{\theta}}^{b_k}_t \\
\delta \dot{\mathbf{b}}_{a_t}        \\
\delta \dot{\mathbf{b}}_{w_t}
\end{bmatrix} 
& \!= \!
\begin{bmatrix}
0 & \mathbf{I} &0 &  0  & 0\\
0 & 0 & -\!\mathbf{R}^{b_k}_t \lfloor \hat{\mathbf{a}}_t - \mathbf{b}_{a_t} \rfloor_{\times} &  -\!\mathbf{R}^{b_k}_t & 0 \\
0 & 0 & -\! \lfloor \hat{\boldsymbol{\omega}}_t-\mathbf{b}_{w_t} \rfloor_{\times} &  0 & -\! \mathbf{I} \\
0 & 0 & 0 & 0 & 0 \\
0 & 0 & 0 & 0 & 0 
\end{bmatrix} 
\!
\begin{bmatrix}
\delta {\boldsymbol{\alpha}}^{b_k}_t \\
\delta {\boldsymbol{\beta}}^{b_k}_t \\
\delta {\boldsymbol{\theta}}^{b_k}_t \\
\delta {\mathbf{b}}_{a_t}        \\
\delta {\mathbf{b}}_{w_t}
\end{bmatrix} \\
&+\!
\begin{bmatrix}
0 & 0 & 0 & 0\\ 
-\mathbf{R}^{b_k}_t & 0 & 0 & 0\\ 
0 & -\mathbf{I} & 0 & 0\\ 
0 & 0 & \mathbf{I} & 0\\ 
0 & 0 & 0 & \mathbf{I}
\end{bmatrix}
\begin{bmatrix}
\mathbf{n}_{a}\\ \mathbf{n}_{w}\\ \mathbf{n}_{b_a}\\ \mathbf{n}_{b_w}
\end{bmatrix} = \mathbf{F}_t \delta\mathbf{z}^{b_k}_t + \mathbf{G}_t \mathbf{n}_t,
\end{aligned}
\end{equation}
$\mathbf{P}^{b_k}_{b_{k+1}}$ can be computed recursively by first-order discrete-time covariance update with the initial covariance $\mathbf{P}^{b_k}_{b_k} = \mathbf{0}$:
\begin{equation}
\begin{aligned}
\mathbf{P}^{b_k}_{t+\delta t} = (\mathbf{I}+\mathbf{F}_t\delta t) \mathbf{P}^{b_k}_t (\mathbf{I}+\mathbf{F}_t\delta t)^T +& (\mathbf{G}_t \delta t) \mathbf{Q} (\mathbf{G}_t \delta t)^T, \\&t \in [k,k+1], 
\end{aligned}
\end{equation}   
where $\mathbf{Q}$ is the diagonal covariance matrix of noise $(\boldsymbol{\sigma}_a^2, \boldsymbol{\sigma}_w^2, \boldsymbol{\sigma}_{b_a}^2, \boldsymbol{\sigma}_{b_w}^2)$.

Meanwhile, the first-order Jacobian matrix $\mathbf{J}_{b_{k+1}}$ of $\delta{\mathbf{z}}^{b_k}_{b_{k+1}}$ with respect to $\delta{\mathbf{z}}^{b_k}_{b_{k}}$ can be also compute recursively with the initial Jacobian $\mathbf{J}_{b_{k}} = \mathbf{I}$,
\begin{equation}
\begin{aligned}
\mathbf{J}_{t+\delta t} &= (\mathbf{I}+\mathbf{F}_t\delta t) \mathbf{J}_{t}, \  \ t \in [k,k+1].
\end{aligned}
\end{equation}

Using this recursive formulation, we get the covariance matrix  $\mathbf{P}^{b_k}_{b_{k+1}}$ and the Jacobian  $\mathbf{J}_{b_{k+1}}$. 
The first order approximation of $\boldsymbol{\alpha}^{b_k}_{b_{k+1}}, \boldsymbol{\beta}^{b_k}_{b_{k+1}}, \boldsymbol{\gamma}^{b_k}_{b_{k+1}}$ with respect to biases can be write as:
\begin{equation}
\label{eq:correct_approximation}
\begin{aligned}
\boldsymbol{\alpha}^{b_k}_{b_{k+1}} &\approx \hat{\boldsymbol{\alpha}}^{b_k}_{b_{k+1}} + \mathbf{J}^\alpha_{b_a} \delta \mathbf{b}_{a_k} + \mathbf{J}^\alpha_{b_w} \delta \mathbf{b}_{w_k} \\
\boldsymbol{{\beta}}^{b_k}_{b_{k+1}} &\approx \hat{\boldsymbol{\beta}}^{b_k}_{b_{k+1}} + \mathbf{J}^\beta_{b_a} \delta \mathbf{b}_{a_k} + \mathbf{J}^\beta_{b_w} \delta \mathbf{b}_{w_k} \\
\boldsymbol{\gamma}^{b_k}_{b_{k+1}} &\approx \hat{\boldsymbol{\gamma}}^{b_k}_{b_{k+1}} \otimes 
\begin{bmatrix}
1\\
\frac{1}{2} \mathbf{J}^\gamma_{b_w} \delta\mathbf{b}_{w_k}
\end{bmatrix}
\end{aligned}
\end{equation}
where $\mathbf{J}^\alpha_{b_a}$ and is the sub-block matrix in $\mathbf{J}_{b_{k+1}}$ whose location is corresponding to $\frac{\delta \boldsymbol{\alpha}^{b_k}_{b_{k+1}}}{\delta \mathbf{b}_{a_k}}$. 
The same meaning is also used for $\mathbf{J}^\alpha_{b_w}, \mathbf{J}^\beta_{b_a}, \mathbf{J}^\beta_{b_w}, \mathbf{J}^\gamma_{b_w}$. 
When the estimation of bias changes slightly, we use~\eqref{eq:correct_approximation} to correct pre-integration results approximately instead of re-propagation.

Now we are able to write down the IMU measurement model with its corresponding covariance ${\mathbf{P}^{b_k}_{b_{k+1}}}$:
\begin{equation}
\label{eq:propagation_function}
\begin{split}
\begin{bmatrix}
\hat{\boldsymbol{\alpha}}^{b_k}_{b_{k+1}}\\
\hat{\boldsymbol{\beta}}^{b_k}_{b_{k+1}}\\
\hat{\boldsymbol{\gamma}}^{b_k}_{b_{k+1}}\\
\mathbf{0}\\
\mathbf{0}\\
\end{bmatrix}
=
\begin{bmatrix}
\mathbf{R}^{b_k}_{w}(\mathbf{p}^{w}_{b_{k+1}} - \mathbf{p}^{w}_{b_k} + \frac{1}{2}\mathbf{g}^{w} \Delta t_k^2 - \mathbf{v}^{w}_{b_k} \Delta t_k)  \\
\mathbf{R}^{b_k}_{w}(  \mathbf{v}^{w}_{b_{k+1}}   + \mathbf{g}^{w} \Delta t_k- \mathbf{v}^{w}_{b_k})  \\
\mathbf{q}^{w^{-1}}_{b_{k}} \otimes\mathbf{q}^{w}_{b_{k+1}}\\
{\mathbf{b}_a}_{b_{k+1}} - {\mathbf{b}_a}_{b_k}\\
{\mathbf{b}_w}_{b_{k+1}} -{\mathbf{b}_w}_{b_k}\\
\end{bmatrix}.
\end{split}
\end{equation}

\section{Estimator Initialization}
\label{sec:Initialization}

Monocular tightly-coupled visual-inertial odometry is a highly nonlinear system.
Since the scale is not directly observable from a monocular camera, it is hard to directly fuse these two measurements without good initial values. 
One may assume a stationary initial condition to start the monocular VINS estimator.
However, this assumption is inappropriate as initialization under motion is frequently encountered in real-world applications.
The situation becomes even more complicated when IMU measurements are corrupted by large bias.
In fact, initialization is usually the most fragile step for monocular VINS. 
A robust initialization procedure is needed to ensure the applicability of the system.

\begin{figure}
    \centering
    \includegraphics[width=0.45\textwidth]{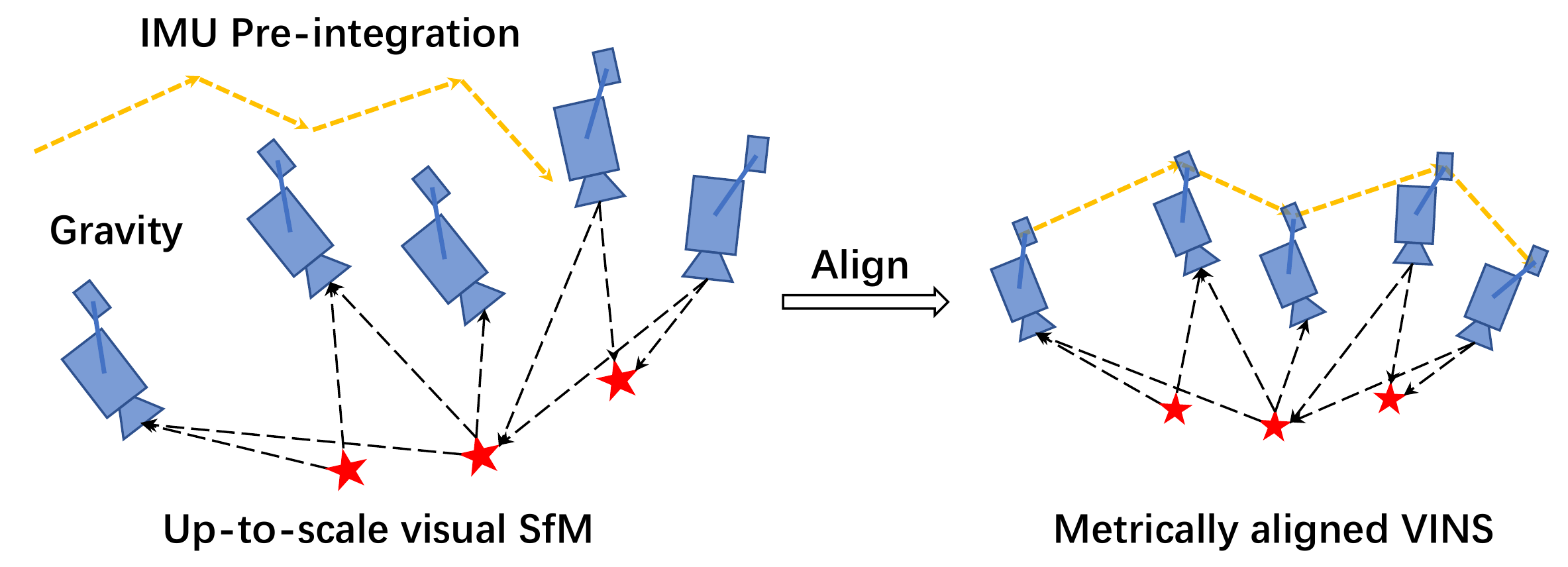}
    \caption{An illustration of the visual-inertial alignment process for estimator initialization.
        \label{fig:align}}
\end{figure}

We adopt a loosely-coupled sensor fusion method to get initial values. 
We find that vision-only SLAM, or Structure from Motion (SfM), has a good property of initialization. 
In most cases, a visual-only system can bootstrap itself by derived initial values from relative motion methods, 
such as the eight-point~\cite{heyden2005multiple} or five-point~\cite{nister2004efficient} algorithms, or estimating the homogeneous matrices.
By aligning metric IMU pre-integration with the visual-only SfM results, we can roughly recover scale, gravity, velocity, and even bias.
This is sufficient for bootstrapping a nonlinear monocular VINS estimator, as shown in Fig.~\ref{fig:align}.

In contrast to~\cite{mur2016visual}, which simultaneously estimates gyroscope and accelerometer bias during the initialization phase,
we choose to ignore accelerometer bias terms in the initial step.
Accelerometer bias is coupled with gravity, and due to the large magnitude of the gravity vector comparing to platform dynamics, and the relatively short during of the initialization phase,
these bias terms are hard to be observed. 
A detailed analysis of accelerometer bias calibration is presented in our previous work~\cite{LiuShen17}.

\subsection{Sliding Window Vision-Only SfM}
\label{subsec:sfm}
The initialization procedure starts with a vision-only SfM to estimate a graph of up-to-scale camera poses and feature positions.

We maintain a sliding window of frames for bounded computational complexity.
Firstly, we check feature correspondences between the latest frame and all previous frames.
If we can find stable feature tracking (more than 30 tracked features) and sufficient parallax (more than 20 rotation-compensated pixels) between the latest frame and any other frames in the sliding window.
we recover the relative rotation and up-to-scale translation between these two frames using the Five-point algorithm~\cite{nister2004efficient}.
Otherwise, we keep the latest frame in the window and wait for new frames. 
If the five-point algorithm success, we arbitrarily set the scale and triangulate all features observed in these two frames. 
Based on these triangulated features, a perspective-n-point (PnP) method~\cite{lepetit2009epnp} is performed to estimate poses of all other frames in the window. 
Finally, a global full bundle adjustment~\cite{triggs1999bundle} is applied to minimize the total reprojection error of all feature observations. 
Since we do not yet have any knowledge about the world frame, we set the first camera frame $(\cdot)^{c_0}$ as the reference frame for SfM.
All frame poses ($\bar{\mathbf{p}}^{c_0}_{c_k}$,$\mathbf{q}^{c_0}_{c_k}$) and feature positions are represented with respect to $(\cdot)^{c_0}$.
Suppose we have a rough measure extrinsic parameters ($ \mathbf{p}^b_c,\mathbf{q}^b_{c}$) between the camera and the IMU, we can translate poses from camera frame to body (IMU) frame,
\begin{equation}
\label{eq:visual_structure}
\begin{split}
\mathbf{q}^{c_0}_{b_k} &= \mathbf{q}^{c_0}_{c_k}  \otimes \inv{(\mathbf{q}^b_c)} \\
s\bar{\mathbf{p}}^{c_0}_{b_k} &= s\bar{\mathbf{p}}^{c_0}_{c_k} - \mathbf{R}^{c_0}_{b_k}\mathbf{p}^b_{c},
\end{split}
\end{equation}
where $s$ the scaling parameter that aligns the visual structure to the metric scale.
Solving this scaling parameter is the key to achieve successful initialization.

\subsection{Visual-Inertial Alignment}
\label{sec:Visual-Inertial Alignment}
\subsubsection{Gyroscope Bias Calibration}
Consider two consecutive frames $b_k$ and $b_{k+1}$ in the window, 
we get the rotation $\mathbf{q}^{c_0}_{b_k}$ and $\mathbf{q}^{c_0}_{b_{k+1}}$ from the visual SfM, 
as well as the relative constraint $\hat{\boldsymbol{\gamma}}^{b_k}_{b_{k+1}}$ from IMU pre-integration. 
We linearize the IMU pre-integration term with respect to gyroscope bias and minimize the following cost function:
\begin{equation}
\label{eq:calib_gyr}
\begin{split}
\min_{\delta b_w} \sum_{k \in \mathcal{B}} \left \| 
\inv{\mathbf{q}^{c_0}_{b_{k+1}}} \otimes \mathbf{q}^{c_0}_{b_{k}} \otimes  \boldsymbol{\gamma}^{b_k}_{b_{k+1}} \right \|^2\\
\boldsymbol{\gamma}^{b_k}_{b_{k+1}} \approx \hat{\boldsymbol{\gamma}}^{b_k}_{b_{k+1}} \otimes 
\begin{bmatrix}
1\\
\frac{1}{2} \mathbf{J}^\gamma_{b_w} \delta\mathbf{b}_w
\end{bmatrix}
,
\end{split}  
\end{equation}
where $\mathcal{B}$ indexes all frames in the window. 
We have the first order approximation of $\hat{\boldsymbol{\gamma}}^{b_k}_{b_{k+1}}$ with respect to the gyroscope bias using the bias Jacobian derived in Sect.~\ref{subsec:imu_pre-integration}. 
In such way, we get an initial calibration of the gyroscope bias $\mathbf{b}_w$.  
Then we re-propagate all IMU pre-integration terms $\hat{\boldsymbol{\alpha}}^{b_k}_{b_{k+1}}, \hat{\boldsymbol{\beta}}^{b_k}_{b_{k+1}}$, and $\hat{\boldsymbol{\gamma}}^{b_k}_{b_{k+1}}$ using the new gyroscope bias.

\subsubsection{Velocity, Gravity Vector and Metric Scale Initialization}
After the gyroscope bias is initialized, we move on to initialize other essential states for navigation, namely velocity, gravity vector, and metric scale:
\begin{equation}
\label{eq:inital_state}
\begin{split}
\mathcal{X}_{I} &= \left [ \mathbf{v}^{b_0}_{b_0},\,\mathbf{v}^{b_1}_{b_1},\, \cdots \,\mathbf{v}^{b_n}_{b_n},\,\mathbf{g}^{c_0},\, s \right ],
\end{split}
\end{equation}
where $\mathbf{v}^{b_k}_{b_k}$ is velocity in the body frame while taking the $k^{th}$ image, $\mathbf{g}^{c_0}$ is the gravity vector in the $c_0$ frame, and $s$ scales the monocular SfM to metric units.

Consider two consecutive frames $b_k$ and $b_{k+1}$ in the window, then~\eqref{eq:local_integration} can be written as:
\begin{equation}
\label{eq:solve_scale_basic}
\begin{split}
{\boldsymbol{\alpha}}^{b_k}_{b_{k+1}} &= \mathbf{R}^{b_k}_{c_0} (s(\bar{\mathbf{p}}^{c_0}_{b_{k+1}}  - \bar{\mathbf{p}}^{c_0}_{b_k}) +  \frac{1}{2} \mathbf{g}^{c_0} {\Delta t}_k^2  - \mathbf{R}^{c_0}_{b_k} \mathbf{v}^{b_k}_{b_k}{\Delta t}_k)
\\
{\boldsymbol{\beta}}^{b_k}_{b_{k+1}} &= \mathbf{R}^{b_k}_{c_0}  (\mathbf{R}^{c_0}_{b_{k+1}}\mathbf{v}^{b_{k+1}}_{b_{k+1}} + \mathbf{g}^{c_0} {\Delta t}_k - \mathbf{R}^{c_0}_{b_k}\mathbf{v}^{b_k}_{b_{k}}).
\end{split}
\end{equation}
We can combine~\eqref{eq:visual_structure} and~\eqref{eq:solve_scale_basic} into the following linear measurement model:
\begin{equation}
\label{eq:linear_eq}
\hat{\mathbf{z}}^{b_k}_{b_{k+1}} =
\begin{bmatrix}
\hat{\boldsymbol{\alpha}}^{b_k}_{b_{k+1}} - \mathbf{p}^b_c + \mathbf{R}^{b_k}_{c_{0}}\mathbf{R}^{c_0}_{b_{k+1}}\mathbf{p}^b_c\\
\hat{\boldsymbol{\beta}}^{b_k}_{b_{k+1}}
\end{bmatrix} 
 = \mathbf{H}^{b_k}_{b_{k+1}} \mathcal{X}_I + \mathbf{n}^{b_k}_{b_{k+1}}  
\end{equation}
where,
\begin{equation}
\mathbf{H}^{b_k}_{b_{k+1}}=
\begin{bmatrix}
 -\mathbf{I}{\Delta t}_k \!&\!\mathbf{0} \!&\!  \frac{1}{2}\mathbf{R}^{b_k}_{c_0}  {\Delta t}_k^2 \!&\! \mathbf{R}^{b_k}_{c_0}(\bar{\mathbf{p}}^{c_0}_{c_{k+1}}-\bar{\mathbf{p}}^{c_0}_{c_k})\\
-\mathbf{I} \!&\!\mathbf{R}^{b_k}_{c_0}\mathbf{R}^{c_0}_{b_{k+1}} \!&\! \mathbf{R}^{b_k}_{c_0} {\Delta t}_k \!&\! \mathbf{0}
\end{bmatrix}\!
\end{equation}
It can be seen that $\mathbf{R}^{c_0}_{b_k},\mathbf{R}^{c_0}_{b_{k+1}},\bar{\mathbf{p}}^{c_0}_{c_{k}}, \bar{\mathbf{p}}^{c_0}_{c_{k+1}}$ are obtained from the up-to-scale monocular visual SfM.
${\Delta t}_k$ is the time interval between two consecutive frames.
By solving this linear least square problem:
\begin{equation}
\label{eq:solve_scale}
\begin{split}
\min_{\mathcal{X}_I}\sum_{k \in \mathcal{B}} \left \|\hat{\mathbf{z}}^{b_k}_{b_{k+1}} -   \mathbf{H}^{b_k}_{b_{k+1}} \mathcal{X}_I\right \|^2,
\end{split}  
\end{equation}
we can get body frame velocities for every frame in the window, the gravity vector in the visual reference frame $(\cdot)^{c_0}$, as well as the scale parameter.

\subsubsection{Gravity Refinement}
\label{subsec:refine_g}

\begin{figure}[t]
    \centering
    \includegraphics[width=0.25\textwidth]{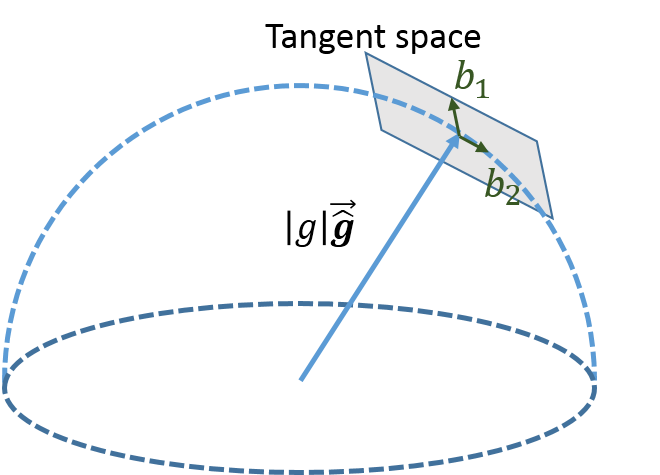}
    \caption{
        Illustration of 2 DOF parameterization of gravity. 
        Since the magnitude of gravity is known, $\mathbf{g}$ lies on a sphere with radius $g \approx 9.81m/s^2$. 
        The gravity is parameterized around current estimate as $g\cdot\hat{\bar{\mathbf{g}}} + w_1\mathbf{b}_1+ w_2\mathbf{b}_2$, 
        where $\mathbf{b}_1$ and $\mathbf{b}_2$ are two orthogonal basis spanning the tangent space.
        \label{fig:gravity}}
\end{figure}

The gravity vector obtained from the previous linear initialization step can be refined by constraining the magnitude.  
In most cases, the magnitude of gravity vector is known.
This results in only 2 DOF remaining for the gravity vector.
We therefore re-parameterize the gravity with two variables on its tangent space.
Our parameterization represents the gravity vector as $g\cdot\bar{\hat{\mathbf{g}}} + w_1\mathbf{b}_1+ w_2\mathbf{b}_2$, where $g$ is the know magnitude of the gravity, $\hat{\bar{\mathbf{g}}}$ is a unit vector representing the gravity direction.
$\mathbf{b}_1$ and $\mathbf{b}_2$ are two orthogonal basis spanning the tangent plane, as shown in Fig.~\ref{fig:gravity}.
$w_1$ and $w_2$ are corresponding displacements towards $\mathbf{b}_1$ and $\mathbf{b}_2$, respectively.
We can find one set of $\mathbf{b}_1$, $\mathbf{b}_2$ by cross products operations using Algorithm~\ref{alg:basis}. 
Then we substitute $\mathbf{g}$ in~\eqref{eq:solve_scale_basic} by $g\cdot\hat{\bar{\mathbf{g}}} + w_1\mathbf{b}_1+ w_2\mathbf{b}_2$, and solve for $w_1$ and $w_2$ together with other state variables.
This process iterates until $\hat{\mathbf{g}}$ converges.

\begin{algorithm}[h]
\label{alg:basis}
    \eIf{$\bar{\hat{\mathbf{g}}} \neq [1, 0, 0]$}{
         $\mathbf{b}_1 \gets normalize(\bar{\hat{\mathbf{g}}} \times [1, 0, 0]) $;
    }{$\mathbf{b}_1 \gets normalize(\bar{\hat{\mathbf{g}}} \times [0, 0, 1]) $;}
    $\mathbf{b}_2 \gets \bar{\hat{\mathbf{g}}} \times \mathbf{b}_1 $;
    \caption{Finding $\mathbf{b}_1$ and $\mathbf{b}_2$}
\end{algorithm}

\subsubsection{Completing Initialization}
After refining the gravity vector, we can get the rotation $\mathbf{q}^w_{c_0}$ between the world frame and the camera frame $c_0$ by rotating the gravity to the z-axis.
We then rotate all variables from reference frame $(\cdot)^{c_0}$ to the world frame $(\cdot)^w$.
The body frame velocities will also be rotated to world frame.
Translational components from the visual SfM will be scaled to metric units.
At this point, the initialization procedure is completed and all these metric values will be fed for a tightly-coupled monocular VIO.

\section{Tightly-coupled Monocular VIO}
\label{sec:Optimization}
After estimator initialization, we proceed with a sliding window-based tightly-coupled monocular VIO for high-accuracy and robust state estimation. 
An illustration of the sliding window formulation is shown in Fig.~\ref{fig:window}. 

\subsection{Formulation}
The full state vector in the sliding window is defined as:
\begin{equation}
\label{eq:variable}
\begin{split}
\mathcal{X}    &= \left [ \mathbf{x}_0,\,\mathbf{x}_{1},\, \cdots \,\mathbf{x}_{n},\, \mathbf{x}^b_c,\, \lambda_0,\,\lambda_{1},\, \cdots \,\lambda_{m} \right ] \\
\mathbf{x}_k   &= \left [ \mathbf{p}^w_{b_k},\,\mathbf{v}^w_{b_k},\,\mathbf{q}^w_{b_k}, \,\mathbf{b}_a, \,\mathbf{b}_g \right ], k\in [0,n] \\
\mathbf{x}^b_c  &= \left [ \mathbf{p}^b_c,\,\mathbf{q}^b_{c} \right ],
\end{split}
\end{equation}
where $\mathbf{x}_k$ is the IMU state at the time that the $k^{th}$ image is captured.
It contains position, velocity, and orientation of the IMU in the world frame, and acceleration bias and gyroscope bias in the IMU body frame. 
$n$ is the total number of keyframes, and $m$ is the total number of features in the sliding window. 
$\lambda_l$ is the inverse depth of the $l^{th}$ feature from its first observation. 

We use a visual-inertial bundle adjustment formulation.
We minimize the sum of prior and the Mahalanobis norm of all measurement residuals to obtain a maximum posteriori estimation:
\begin{equation}
\label{eq:nonlinear_cost_function}
\begin{aligned}
\min_{\mathcal{X}} \left \{ \left 
\| \mathbf{r}_p - \mathbf{H}_p \mathcal{X} \right \|^2 
+ \sum_{k \in \mathcal{B}}  \left \| \mathbf{r}_{\mathcal{B}}(\hat{\mathbf{z}}^{b_k}_{b_{k+1}},\, \mathcal{X}) \right \|_{\mathbf{P}^{b_k}_{b_{k+1}}}^2 + \right. \\
\left. 
\sum_{(l,j) \in \mathcal{C}} \rho( \left \| \mathbf{r}_{\mathcal{C}}(\hat{\mathbf{z}}^{c_j}_l ,\, \mathcal{X}) \right \|_{\mathbf{P}^{c_j}_l}^2 )
 \right\},
\end{aligned}                    
\end{equation}
where the Huber norm \cite{Hub64} is defined as:
\begin{equation}
\rho(s)=\begin{cases}
1 & s \geq 1,\\
2\sqrt{s}-1 & s < 1.
\end{cases}
\end{equation}
$\mathbf{r}_{\mathcal{B}}(\hat{\mathbf{z}}^{b_k}_{b_{k+1}},\, \mathcal{X})$ and $\mathbf{r}_{\mathcal{C}}(\hat{\mathbf{z}}^{c_j}_l,\, \mathcal{X})$ are residuals for IMU and visual measurements respectively.
Detailed definition of the residual terms will be presented in Sect.~\ref{subsec:imu_measurement_model} and Sect.~\ref{subsec:visual_measurement_model}.
$\mathcal{B}$ is the set of all IMU measurements, $\mathcal{C}$ is the set of features which have been observed at least twice in the current sliding window.
$\{\mathbf{r}_p,\,\mathbf{H}_p\}$ is the prior information from marginalization.
Ceres Solver~\cite{ceres-solver} is used for solving this nonlinear problem.

\subsection{IMU Measurement Residual}
\label{subsec:imu_measurement_model}
Consider the IMU measurements within two consecutive frames $b_k$ and $b_{k+1}$ in the sliding window, 
according to the IMU measurement model defined in~\eqref{eq:propagation_function}, the residual for pre-integrated IMU measurement can be defined as:
\begin{equation}
\begin{split}
& \mathbf{r}_{\mathcal{B}}(\hat{\mathbf{z}}^{b_k}_{b_{k+1}},\, \mathcal{X})=
\begin{bmatrix}
\delta\boldsymbol{\alpha}^{b_k}_{b_{k+1}}\\
\delta\boldsymbol{\beta}^{b_k}_{b_{k+1}}\\
\delta\boldsymbol{\theta}^{b_k}_{b_{k+1}}\\
\delta{\mathbf{b}_a}\\
\delta{\mathbf{b}_g}\\
\end{bmatrix}\\
&=\begin{bmatrix}
\mathbf{R}^{b_k}_{w}(\mathbf{p}^{w}_{b_{k+1}} - \mathbf{p}^{w}_{b_k} + \frac{1}{2}\mathbf{g}^{w} \Delta t_k^2 - \mathbf{v}^{w}_{b_k} \Delta t_k) -\boldsymbol{\hat{\alpha}}^{b_k}_{b_{k+1}} \\
\mathbf{R}^{b_k}_{w}(  \mathbf{v}^{w}_{b_{k+1}}   + \mathbf{g}^{w} \Delta t_k- \mathbf{v}^{w}_{b_k})-  \boldsymbol{\hat{\beta}}^{b_k}_{b_{k+1}} \\
2\begin{bmatrix}\mathbf{q}^{w^{-1}}_{b_{k}} \otimes\mathbf{q}^{w}_{b_{k+1}}
\otimes \inv{(\hat{\boldsymbol{\gamma}}^{b_k}_{b_{k+1}})}
\end{bmatrix}_{xyz}\\
{\mathbf{b}_a}_{b_{k+1}} - {\mathbf{b}_a}_{b_k}\\
{\mathbf{b}_w}_{b_{k+1}} - {\mathbf{b}_w}_{b_k}
\end{bmatrix},
\end{split}
\end{equation} 
where $\begin{bmatrix}\cdot \end{bmatrix}_{xyz}$ extracts the vector part of a quaternion $\mathbf{q}$ for error state representation.
$\delta\boldsymbol{\theta}^{b_k}_{b_{k+1}}$ is the three dimensional error-state representation of quaternion.
$[\boldsymbol{\hat{\alpha}}^{b_k}_{b_{k+1}} ,\, \boldsymbol{\hat{\beta}}^{b_k}_{b_{k+1}} ,\, \hat{\boldsymbol{\gamma}}^{b_k}_{b_{k+1}}]\transpose$ 
are pre-integrated IMU measurement terms using only noisy accelerometer and gyroscope measurements within the time interval between two consecutive image frames.
Accelerometer and gyroscope biases are also included in the residual terms for online correction.

\subsection{Visual Measurement Residual}
\label{subsec:visual_measurement_model}
\begin{figure}
    \centering
    \includegraphics[width=0.3\textwidth]{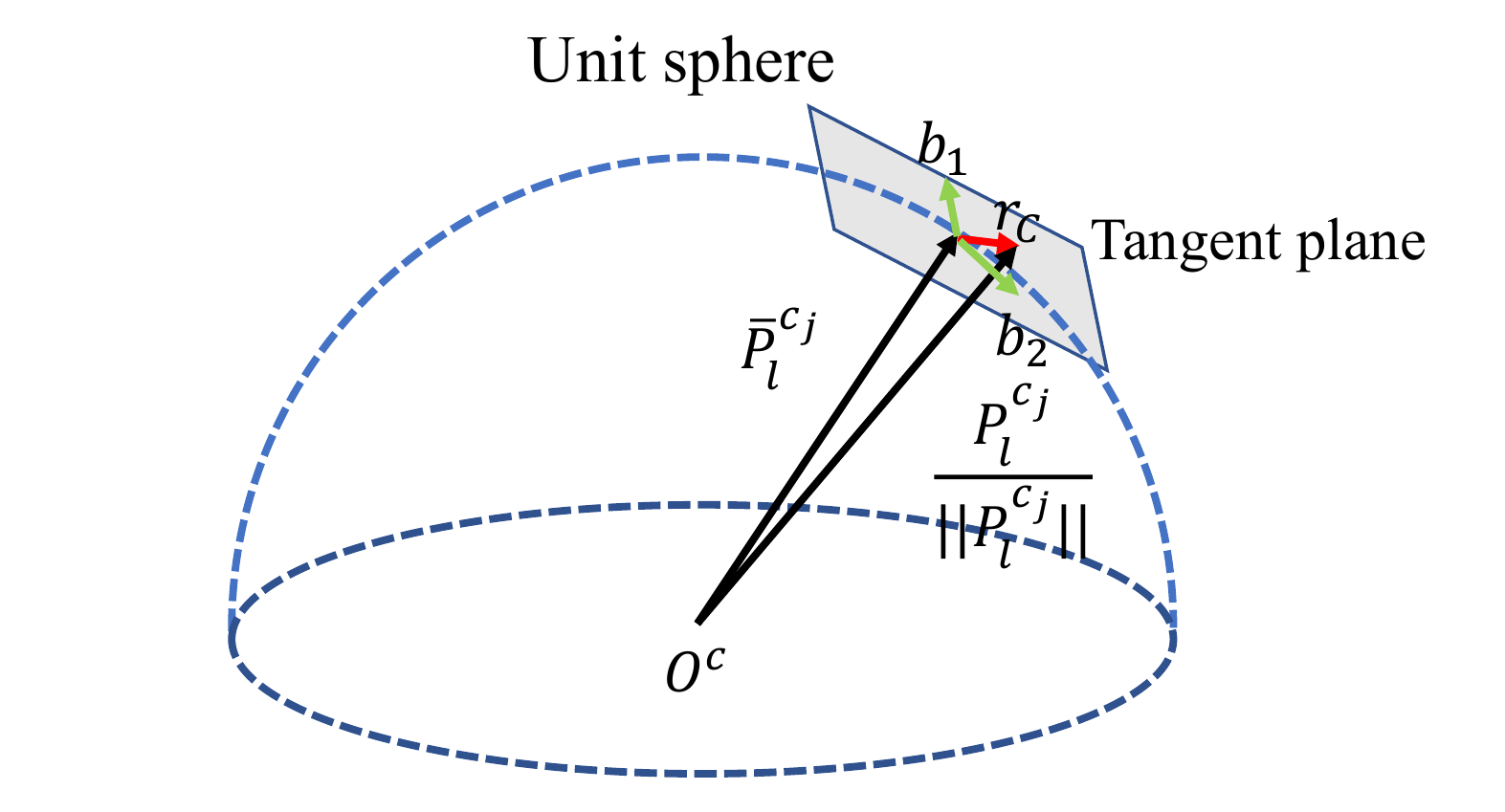}
    \caption{An illustration of the visual residual on a unit sphere. 
             $\hat{\bar{\mathcal{P}}}^{c_j}_l$ is the unit vector for the observation of the $l^{th}$ feature in the $j^{th}$ frame.
             $\mathcal{P}^{c_j}_l$ is predicted feature measurement on the unit sphere by transforming its first observation in the $i^{th}$ frame to the $j^{th}$ frame.
             The residual is defined on the tangent plane of $\hat{\bar{\mathcal{P}}}^{c_j}_l$.
        \label{fig:sphere_model}}
\end{figure}

In contrast to traditional pinhole camera models that define reprojection errors on a generalized image plane, we define the camera measurement residual on a unit sphere.
The optics for almost all types of cameras, including wide-angle, fisheye or omnidirectional cameras, can be modeled as a unit ray connecting the surface of a unit sphere.
Consider the $l^{th}$ feature that is first observed in the $i^{th}$ image, the residual for the feature observation in the $j^{th}$ image is defined as:
\begin{equation}
\label{eq:vision model}
\begin{split}
&\mathbf{r}_{\mathcal{C}}(\hat{\mathbf{z}}^{c_j}_l,\, \mathcal{X}) =
\begin{bmatrix}
\mathbf{b}_1 \ \ \mathbf{b}_2
\end{bmatrix}^T  
\cdot ({\hat{\bar{\mathcal{P}}}^{c_j}_l} - 
\frac{\mathcal{P}^{c_j}_l}{\| \mathcal{P}^{c_j}_l\|})
\\
&\hat{\bar{\mathcal{P}}}^{c_j}_l =  \inv{\pi_c} (
\begin{bmatrix}
\hat{u}^{c_j}_l \\
\hat{v}^{c_j}_l 
\end{bmatrix}
)\\
&\mathcal{P}^{c_j}_l = 
\mathbf{R}^c_b 
(\mathbf{R}^{b_j}_w
(\mathbf{R}^w_{b_i}
(\mathbf{R}^b_c 
\frac{1}{\lambda_l} 
\inv{\pi_c} (
\begin{bmatrix}
u^{c_i}_l \\
v^{c_i}_l 
\end{bmatrix}
)\\
& \qquad \qquad \qquad \qquad \qquad  + \mathbf{p}^b_c) + \mathbf{p}^w_{b_i} - \mathbf{p}^w_{b_j}) - \mathbf{p}^b_c)
, 
\end{split}  
\end{equation}
where $[ u^{c_i}_l ,\, v^{c_i}_l ] $ is the first observation of the $l^{th}$ feature that happens in the $i^{th}$ image. 
$[ \hat{u}^{c_j}_l ,\, \hat{v}^{c_j}_l ] $ is the observation of the same feature in the $j^{th}$ image. 
${\pi}^{-1}_c$ is the back projection function which turns a pixel location into a unit vector using camera intrinsic parameters. 
Since the degrees-of-freedom of the vision residual is two, we project the residual vector onto the tangent plane. 
$\mathbf{b}_1, \mathbf{b}_2$ are two arbitrarily selected orthogonal bases that span the tangent plane of $\hat{\bar{\mathcal{P}}}^{c_j}_l$, as shown in Fig.~\ref{fig:sphere_model}. 
We can find one set of $\mathbf{b}_1, \mathbf{b}_2$ easily, as shown in Algorithm 1. 
$\mathbf{P}^{c_j}_l$, as used in~\eqref{eq:nonlinear_cost_function}, is the standard covariance of a fixed length in the tangent space.

\begin{figure}
    \centering
    \includegraphics[width=0.5\textwidth]{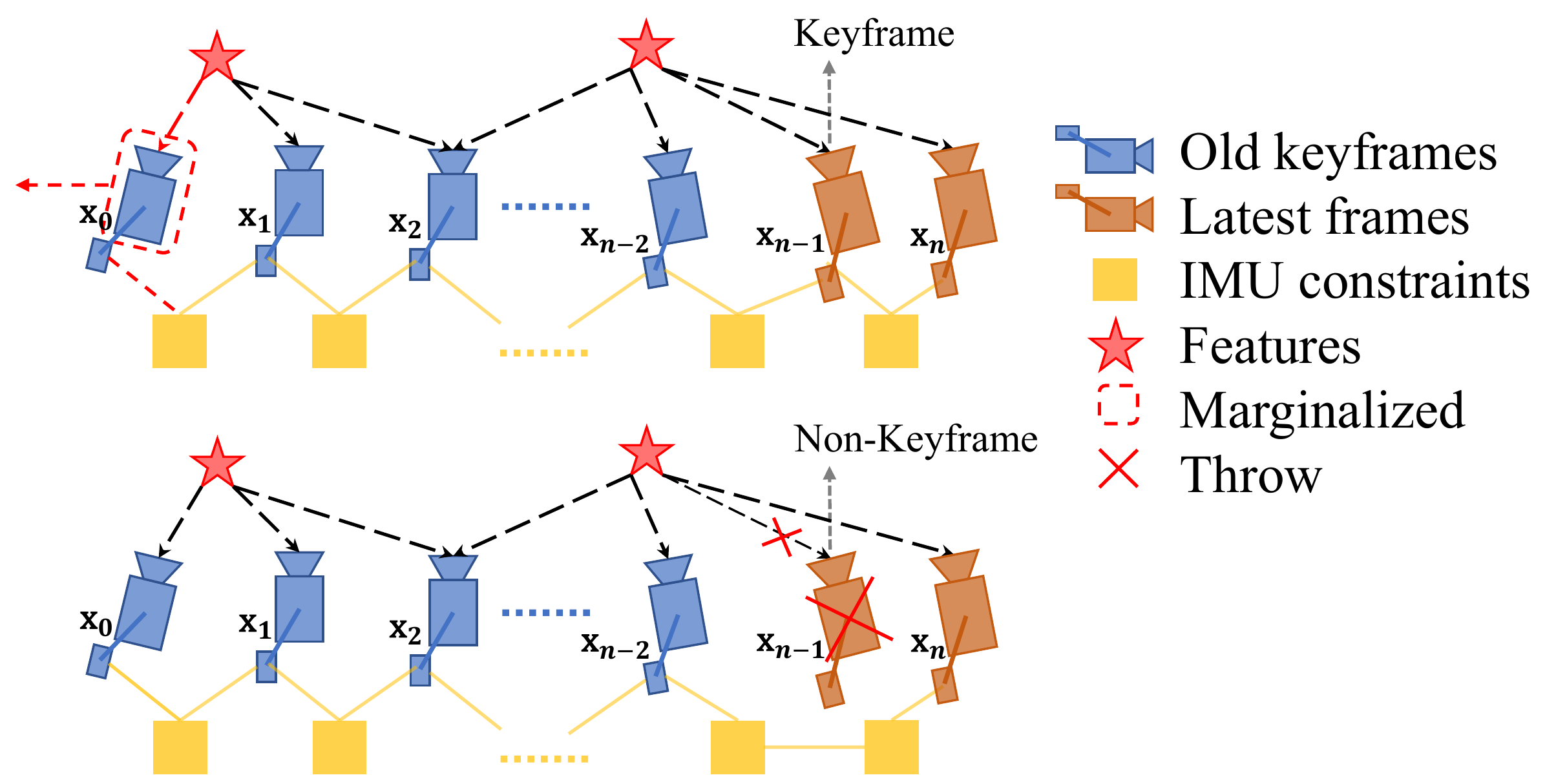}
    \caption{An illustration of our marginalization strategy. 
             If the second latest frame is a keyframe, we will keep it in the window, and marginalize the oldest frame and its corresponding visual and inertial measurements. 
             Marginalized measurements are turned into a prior. 
             If the second latest frame is not a keyframe, we will simply remove the frame and all its corresponding visual measurements.
             However, pre-integrated inertial measurements are kept for non-keyframes, and the pre-integration process is continued towards the next frame.
        \label{fig:marginlization}}
\end{figure}

\subsection{Marginalization}
\label{subsubsection:Marginalization}
In order to bound the computational complexity of our optimization-based VIO, marginalization is incorporated.
We selectively marginalize out IMU states $\mathbf{x}_k$  and features $\lambda_l$ from the sliding window, meanwhile convert measurements corresponding to marginalized states into a prior.

As shown in the Fig.~\ref{fig:marginlization}, when the second latest frame is a keyframe, it will stay in the window, and the oldest frame is marginalized out with its corresponding measurements.
Otherwise, if the second latest frame is a non-keyframe, we throw visual measurements and keep IMU measurements that connect to this non-keyframe.
We do not marginalize out all measurements for non-keyframes in order to maintain sparsity of the system.
Our marginalization scheme aims to keep spatially separated keyframes in the window.
This ensures sufficient parallax for feature triangulation, and maximize the probability of maintaining accelerometer measurements with large excitation.

The marginalization is carried out using the Schur complement \cite{SibMatSuk1009}.
We construct a new prior based on all marginalized measurements related to the removed state.
The new prior is added onto the existing prior.

We do note that marginalization results in the early fix of linearization points, which may result in suboptimal estimation results.
However, since small drifting is acceptable for VIO, we argue that the negative impact caused by marginalization is not critical.

\begin{figure}
    \centering
    \includegraphics[width=0.49\textwidth]{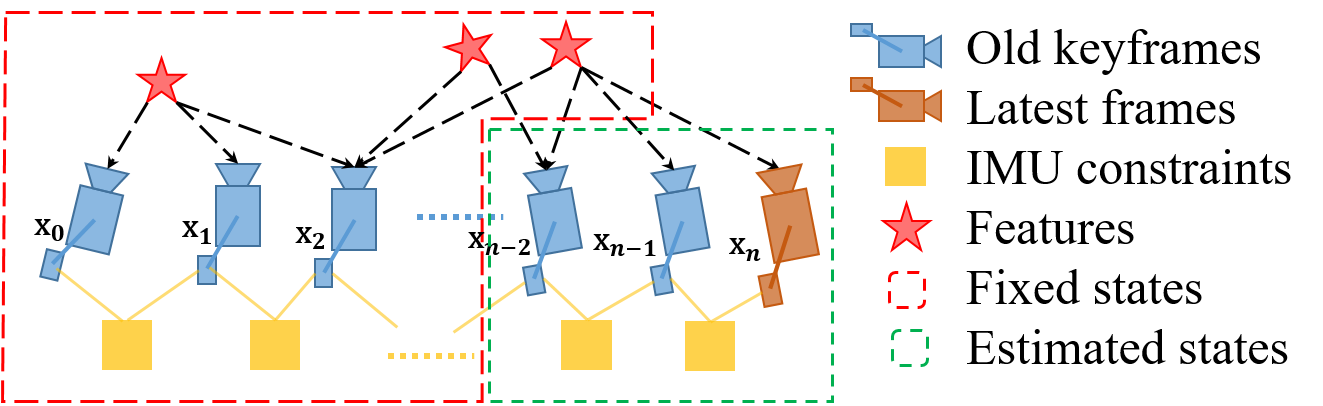}
    \caption{An illustration of motion-only bundle adjustment for camera-rate outputs.
        \label{fig:motionBA}}
\end{figure}

\subsection{Motion-only Visual-Inertial Bundle Adjustment for Camera-Rate State Estimation}

For devices with low computational power, such as mobile phones, the tightly-coupled monocular VIO cannot achieve camera-rate outputs due to the heavy computation demands for the nonlinear optimization. 
To this end, we employ a lightweight motion-only visual-inertial bundle adjustment to boost the state estimation to camera-rate ($~30\,\text{Hz}$).

The cost function for the motion-only visual-inertial bundle adjustment is the same as the one for monocular VIO in~\eqref{eq:nonlinear_cost_function}.
However, instead of optimizing all states in the sliding window, we only optimize the poses and velocities of a fixed number of latest IMU states.
We treat feature depth, extrinsic parameters, bias, and old IMU states that we do not want to optimize as constant values.
We do use all visual and inertial measurements for the motion-only bundle adjustment. 
This results in much smoother state estimates than single frame PnP methods.
An illustration of the proposed strategy is shown in Fig.~\ref{fig:motionBA}.
In contrast to the full tightly-coupled monocular VIO, which may cause more than 50ms on state-of-the-art embedded computers, 
the motion-only visual-inertial bundle adjustment only takes about 5ms to compute.
This enables the low-latency camera-rate pose estimation that is particularly beneficial for drone and AR applications. 

\subsection{IMU Forward Propagation for IMU-Rate State Estimation}
\label{subsec:imu_propagation}
IMU measurements come at a much higher rate than visual measurements.
Although the frequency of our VIO is limited by image capture frequency,
we can still directly propagate the latest VIO estimate with the set of most recent IMU measurements to achieve IMU-rate performance.
The high-frequency state estimates can be utilized as state feedback for closed loop closure.
An autonomous flight experiment utilizing this IMU-rate state estimates is presented in Sect.~\ref{sec:onboard flight}.

\subsection{Failure Detection and Recovery}
Although our tightly-coupled monocular VIO is robust to various challenging environments and motions.
Failure is still unavoidable due to violent illumination change or severely aggressive motions. 
Active failure detection and recovery strategy can improve the practicability of proposed system.
Failure detection is an independent module that detects unusual output from the estimator.
We are currently using the following criteria for failure detection:
\begin{itemize}
    \item The number of features being tracking in the latest frame is less than a certain threshold;
    \item Large discontinuity in position or rotation between last two estimator outputs;
    \item Large change in bias or extrinsic parameters estimation;
\end{itemize}

Once a failure is detected, the system switches back to the initialization phase.
Once the monocular VIO is successfully initialized, a new and separate segment of the pose graph will be created.

\section{Relocalization}
\label{sec:rolocalization}
\begin{figure}
    \centering
         \subfigure[Relocalization]{
         \label{fig:relocalize_system}       
        \includegraphics[width=0.98\columnwidth]{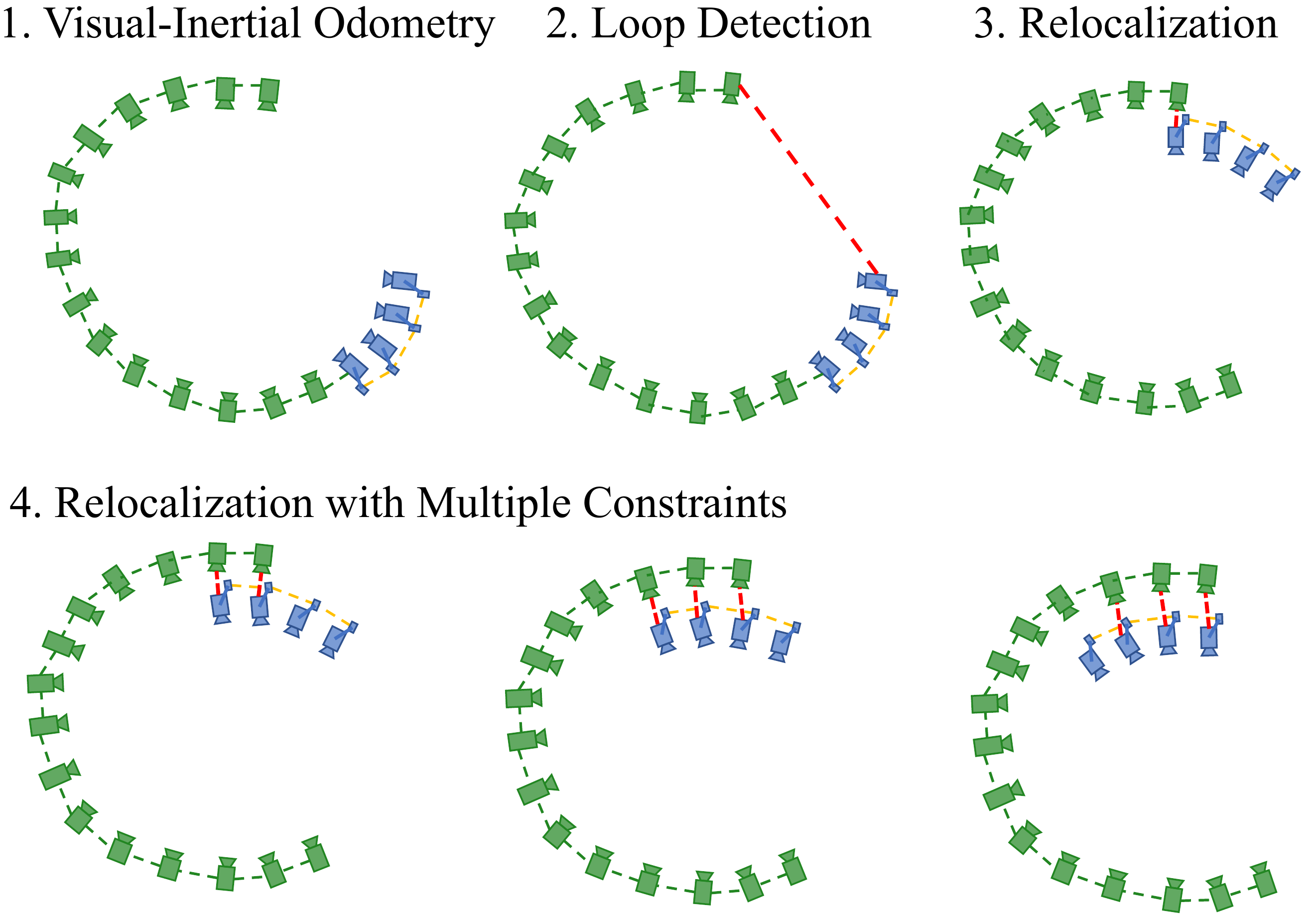}}
         \subfigure[Global Pose Graph optimization]{
         \label{fig:pose_graph_system}
        \includegraphics[width=0.98\columnwidth]{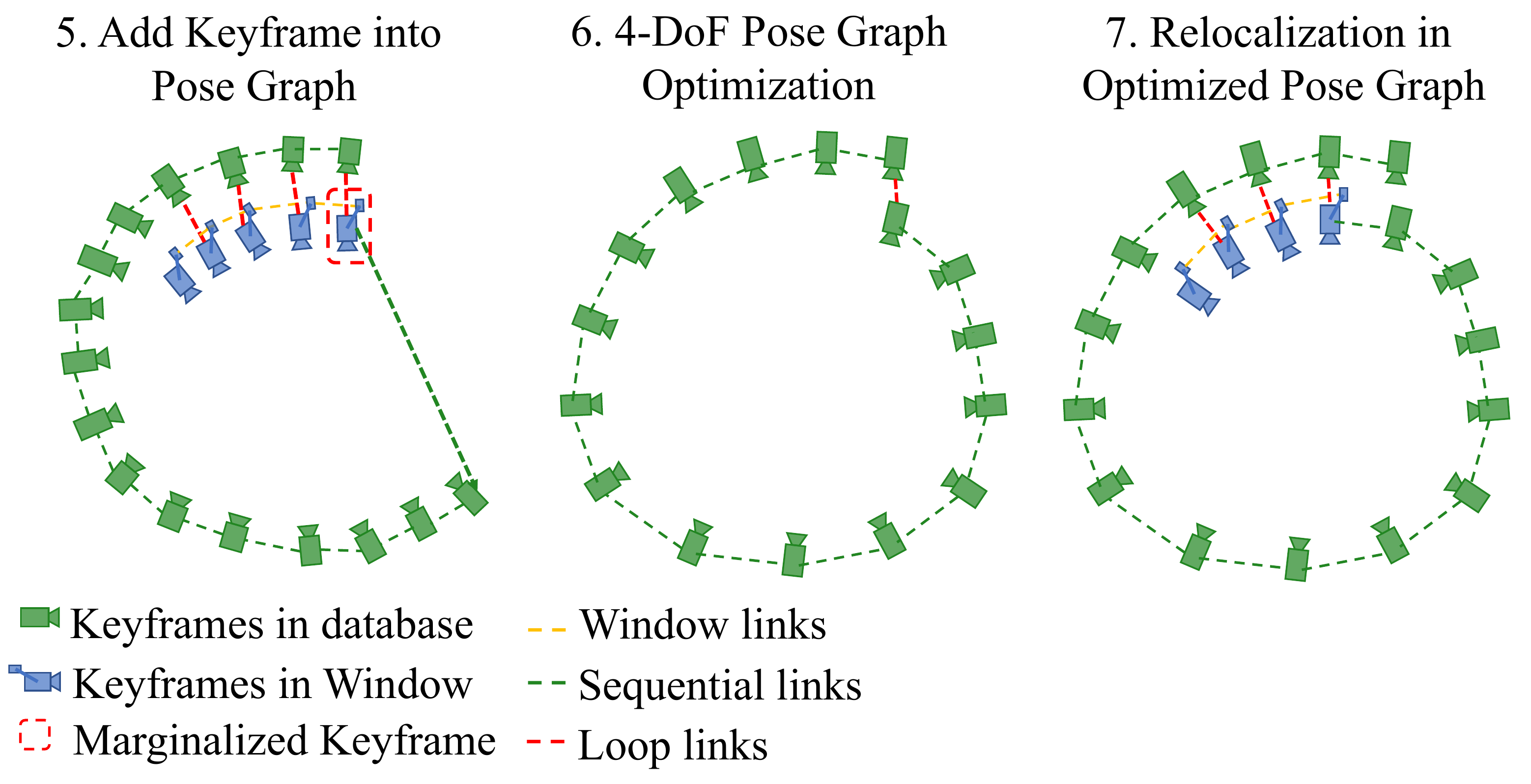}}
    \caption{A diagram illustrating the relocalization and pose graph optimization procedure. 
             Fig.~\ref{fig:relocalize_system} shows the relocalization procedure. 
             It starts with VIO-only pose estimates (blue). Past states are recorded (green). 
             If a loop is detected for the newest keyframe (Sect.~\ref{sec:Loop Detection}), as shown by the red line in the second plot, a relocalization occurred.
             Note that due to the use of feature-level correspondences for relocalization, we are able to incorporate loop closure constraints from multiple past keyframes (Sect.~\ref{sec:relocalization}), as indicated in the last three plots.
             The pose graph optimization is illustrated in Fig.~\ref{fig:pose_graph_system}. 
             A keyframe is added into the pose graph when it is marginalized out from the sliding window. 
             If there is a loop between this keyframe and any other past keyframes, the loop closure constraints, formulated as 4-DOF relative rigid body transforms, will also be added to the pose graph.
             The pose graph is optimized using all relative pose constraints (Sect.~\ref{sec:pose_graph}) in a separate thread, and the relocalization module always runs with respect to the newest pose graph configuration.}
    \label{fig:loop_system}
\end{figure}

Our sliding window and marginalization scheme bound the computation complexity, but it also introduces accumulated drifts for the system.
To be more specific, drifts occur in global 3D position (x, y, z) and the rotation around the gravity direction (yaw).
To eliminate drifts, a tightly-coupled relocalization module that seamlessly integrates with the monocular VIO is proposed.
The relocalization process starts with a loop detection module that identifies places that have already been visited.
Feature-level connections between loop closure candidates and the current frame are then established.
These feature correspondences are tightly integrated into the monocular VIO module, resulting in drift-free state estimates with minimum computation overhead.
Multiple observations of multiple features are directly used for relocalization, resulting in higher accuracy and better state estimation smoothness.
A graphical illustration of the relocalization procedure is shown in Fig.~\ref{fig:relocalize_system}.

\subsection{Loop Detection}
\label{sec:Loop Detection}
We utilize DBoW2~\cite{GalvezTRO12}, a state-of-the-art bag-of-word place recognition approach, for loop detection.
In addition to the corner features that are used for the monocular VIO, 
500 more corners are detected and described by the BRIEF descriptor~\cite{calonder2010brief}.
The additional corner features are used to achieve better recall rate on loop detection.
Descriptors are treated as the visual word to query the visual database.
DBoW2 returns loop closure candidates after temporal and geometrical consistency check.
We keep all BRIEF descriptors for feature retrieving, but discard the raw image to reduce memory consumption. 

We note that our monocular VIO is able to render roll and pitch angles observable. 
As such, we do not need to rely on rotation-invariant features, such as the ORB feature used in ORB SLAM~\cite{mur2015orb}.

\begin{figure}
    \centering
    \subfigure[BRIEF descriptor matching results]{
        \label{fig:matching1}    
        \includegraphics[width=0.9\columnwidth]{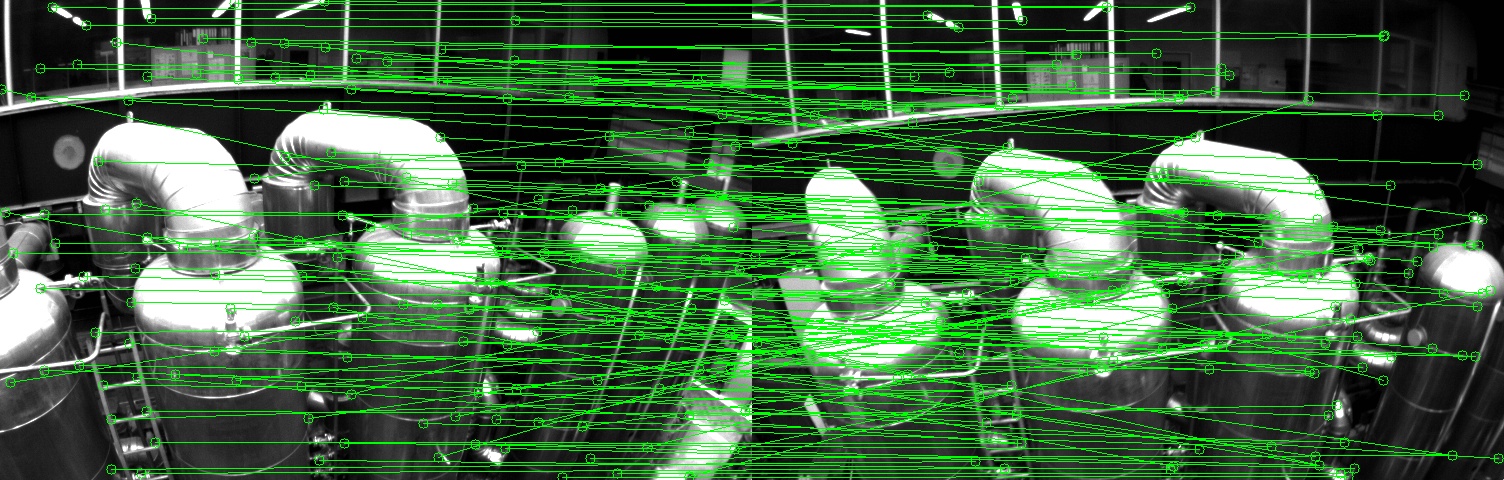}}   
    \subfigure[First step: 2D-2D outlier rejection results]{
        \label{fig:matching0}
        \includegraphics[width=0.9\columnwidth]{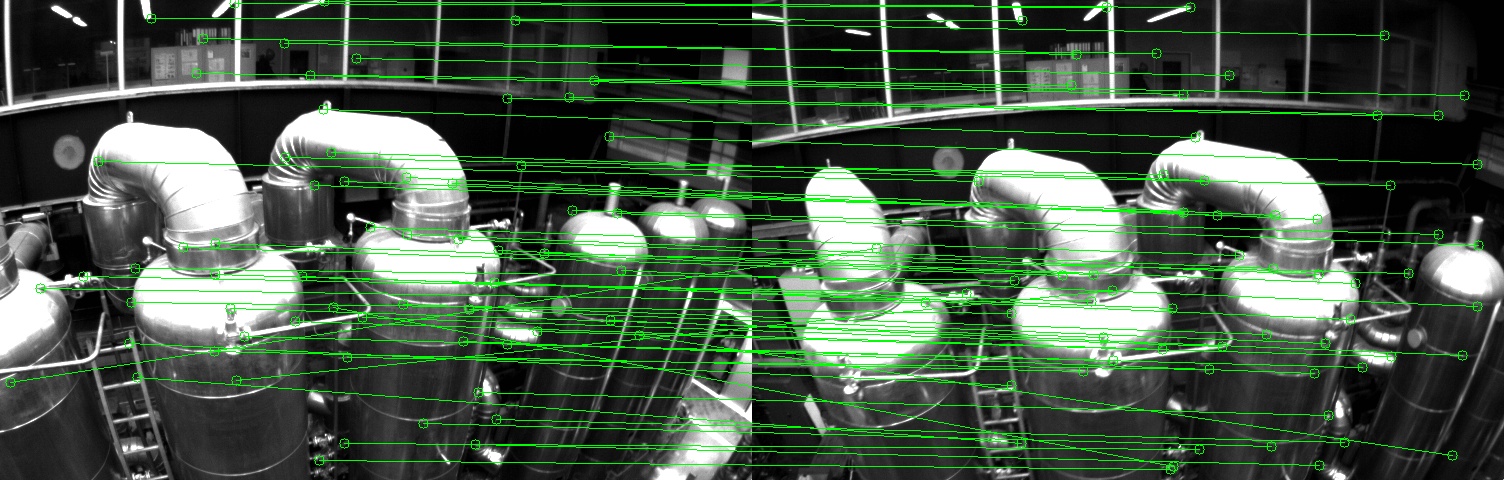}}
    \subfigure[Second step: 3D-2D outlier rejection results.]{
        \label{fig:matching3}
        \includegraphics[width=0.9\columnwidth]{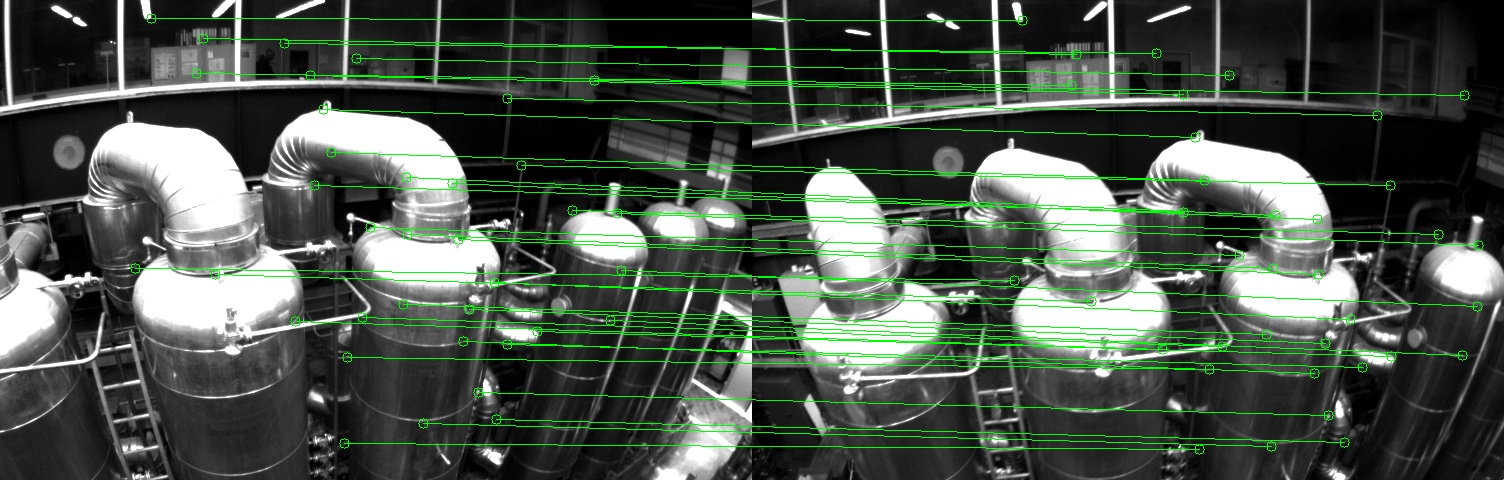}}
    \caption{Descriptor matching and outlier removal for feature retrieval during loop closure.}
    \label{fig:matching}
        \vspace{-0.5cm}
\end{figure}

\subsection{Feature Retrieval}
When a loop is detected, the connection between the local sliding window and the loop closure candidate is established by retrieving feature correspondences.
Correspondences are found by BRIEF descriptor matching.
Directly descriptor matching may cause a large number a lot of outliers.
To this end, we use two-step geometric outlier rejection, as illustrated in Fig.~\ref{fig:matching}.
\begin{itemize}
    \item 2D-2D: fundamental matrix test with RANSAC~\cite{hartley2003multiple}. We use 2D observations of retrieved features in the current image and loop closure candidate image to perform fundamental matrix test.
    \item 3D-2D: PnP test with RANSAC~\cite{lepetit2009epnp}. Based on the known 3D position of features in the local sliding window, and 2D observations in the loop closure candidate image, we perform PnP test. 
\end{itemize}
When the number of inliers beyond a certain threshold, we treat this candidate as a correct loop detection and perform relocalization.

\subsection{Tightly-Coupled Relocalization}
\label{sec:relocalization}

The relocalization process effectively aligns the current sliding window maintained by the monocular VIO (Sect.~\ref{sec:Optimization}) to the graph of past poses.
During relocalization, we treat poses of all loop closure frames as constants.
We jointly optimize the sliding window using all IMU measurements, local visual measurement measurements, and retrieved feature correspondences from loop closure.
We can easily write the visual measurement model for retrieved features observed by a loop closure frame $v$ to be the same as those for visual measurements in VIO, as shown in~\eqref{eq:vision model}. 
The only difference is that the pose ($\hat{\mathbf{q}}^w_v, \hat{\mathbf{p}}^w_v$) of the loop closure frame, 
which is taken from the pose graph (Sect.~\ref{sec:pose graph optimization}), 
or directly from past odometry output (if this is the first relocalization), 
is treated as a constant.
To this end, we can slightly modify the nonlinear cost function in~\eqref{eq:nonlinear_cost_function} with additional loop terms:
\begin{equation}
\begin{aligned}
\label{eq:jointly optimize loop}
\min_{\mathcal{X}}  \left\{ \left\| \mathbf{r}_p - \mathbf{H}_p \mathcal{X} \right\|^2
+ \sum_{k \in \mathcal{B}}  \left\| \mathbf{r}_{\mathcal{B}}(\hat{\mathbf{z}}^{b_k}_{b_{k+1}}, \mathcal{X})  \right\|_{\mathbf{P}^{b_k}_{b_{k+1}}}^2 \right. \\
\left. 
+ \sum_{(l,j) \in \mathcal{C}} \rho ( \left\| \mathbf{r}_{\mathcal{C}}(\hat{\mathbf{z}}^{c_j}_l, \mathcal{X})  \right\|_{\mathbf{P}^{c_j}_l}^2 ) \right. \\
\left. 
+ \sum_{(l,v) \in \mathcal{L}} \rho ( \left\| \mathbf{r}_{\mathcal{C}}(\hat{\mathbf{z}}^{v}_l, \mathcal{X}, \hat{\mathbf{q}}^w_v, \hat{\mathbf{p}}^w_v) \right\|_{\mathbf{P}^{c_v}_l}^2 
\right\},
\end{aligned}                    
\end{equation}
where $\mathcal{L}$ is the set of the observation of retrieved features in the loop closure frames. 
${(l,v)}$ means $l^{th}$ feature observed in the loop closure frame $v$. 
Note that although the cost function is slightly different from~\eqref{eq:nonlinear_cost_function}, the dimension of the states to be solved remains the same, as poses of loop closure frames are considered as constants.
When multiple loop closures are established with the current sliding window, we optimize using all loop closure feature correspondences from all frames at the same time.
This gives multi-view constraints for relocalization, resulting in higher accuracy and better smoothness.
Note that the global optimization of past poses and loop closure frames happens {\it after} relocalization, and will be discussed in Sect.~\ref{sec:pose graph optimization}.

\section{Global Pose Graph Optimization}
\label{sec:pose graph optimization}
After relocalization, the local sliding window shifts and aligns with past poses.
Utilizing the relocalization results, this additional pose graph optimization step is developed to ensure the set of past poses are registered into a globally consistent configuration.

Since our visual-inertial setup renders roll and pitch angles fully observable, the accumulated drift only occurs in four degrees-of-freedom (x, y, z and yaw angle). 
To this end, we ignore estimating the drift-free roll and pitch states, and only perform 4-DOF pose graph optimization.

\subsection{Adding Keyframes into the Pose Graph}
\label{sec:Adding Keyframe into Pose Graph}
When a keyframe is marginalized out from the sliding window, it will be added to pose graph.
This keyframe serves as a vertex in the pose graph, and it connects with other vertexes by two types of edges:

\subsubsection{Sequential Edge} 
a keyframe will establish several sequential edges to its previous keyframes. 
A sequential edge represents the relative transformation between two keyframes in the local sliding window, which value is taken directly from VIO.
Considering a newly marginalized keyframe $i$ and one of its previous keyframes $j$, the sequential edge only contains relative position $\hat{\mathbf{p}}^i_{ij}$ and yaw angle $\hat{\psi}_{ij}$.
\begin{equation}
\label{eq:loop constraint}
\begin{aligned}
\hat{\mathbf{p}}^i_{ij} =& \inv{\hat{\mathbf{R}}^w_i}(\hat{\mathbf{p}}^w_j - \hat{\mathbf{p}}^w_i)\\
\hat{\psi}_{ij} =& \hat{{\psi}}_{j} -  \hat{\psi}_{i}.
\end{aligned}                    
\end{equation}

\subsubsection{Loop Closure Edge} 
If the newly marginalized keyframe has a loop connection, it will be connected with the loop closure frame by a loop closure edge in the pose graph. 
Similarly, the loop closure edge only contains 4-DOF relative pose transform that is defined the same as~\eqref{eq:loop constraint}.
The value of the loop closure edge is obtained using results from relocalization.

\subsection{4-DOF Pose Graph Optimization}
\label{sec:pose_graph}

We define the residual of the edge between frames $i$ and $j$ minimally as:  
\begin{equation}
\label{eq:pose graph residual}
\begin{split}
\mathbf{r}_{i,j}(\mathbf{p}^w_i, \psi_i, \mathbf{p}^w_j, \psi_j)=
\begin{bmatrix}
\inv{\mathbf{R}(\hat{\phi}_i,\hat{\theta}_i,\psi_i)} (\mathbf{p}^w_j - \mathbf{p}^w_i)-\hat{\mathbf{p}}^i_{ij}\\
\psi_j - \psi_i - \hat{\psi}_{ij}\\
\end{bmatrix}
\end{split},
\end{equation} 
where $\hat{\phi}_i,\hat{\theta}_i$ are the estimates roll and pitch angles, which are obtained directly from monocular VIO. 

The whole graph of sequential edges and loop closure edges are optimized by minimizing the following cost function:
\begin{equation}
\label{eq:pose graph optimization}
\begin{aligned}
\min_{\mathbf{p}, \psi} \left \{ 
 \sum_{(i,j) \in \mathcal{S}} \left \| \mathbf{r}_{i,j} \right \|^2  + 
 \sum_{(i,j) \in \mathcal{L}}   \rho (\left  \| \mathbf{r}_{i,j} \right \|^2 ) \right\},
\end{aligned}                    
\end{equation}
where $\mathcal{S}$ is the set of all sequential edges and $\mathcal{L}$ is the set of all loop closure edges. 
Although the tightly-coupled relocalization already helps with eliminating wrong loop closures,
we add another Huber norm $\rho(\cdot)$ to further reduce the impact of any possible wrong loops.
In contrast, we do not use any robust norms for sequential edges, as these edges are extracted from VIO, which already contain sufficient outlier rejection mechanisms.

Pose graph optimization and relocalization (Sect.~\ref{sec:relocalization}) runs asynchronously in two separate threads. 
This enables immediate use of the most optimized pose graph for relocalization whenever it becomes available.
Similarly, even if the current pose graph optimization is not completed yet, relocalization can still take place using the existing pose graph configuration.
This process is illustrated in Fig.~ \ref{fig:pose_graph_system}.

\subsection{Pose Graph Management}
The size of the pose graph may grow unbounded when the travel distance increases, limiting the real-time performance of the system in the long run.
To this end, we implement a downsample process to maintain the pose graph database to a limited size.
All keyframes with loop closure constraints will be kept, while other keyframes that are either too close or have very similar orientations to its neighbors may be removed.
The probability of a keyframe being removed is proportional to its spatial density to its neighbors. 

\section{Experimental Results}
\label{sec:experiment}

We perform three experiments and two applications to evaluate the proposed VINS-Mono system. 
In the first experiment, we compare the proposed algorithm with another state-of-the-art algorithm on public datasets. 
We perform a numerical analysis to show the accuracy of our system.
We then test our system in the indoor environment to evaluate the performance in repetitive scenes.
A large-scale experiment is carried out to illustrate the long-time practicability of our system.
Additionally, we apply the proposed system for two applications. 
For aerial robot application, we use VINS-Mono for position feedback to control a drone to follow a pre-defined trajectory. 
We then port our approach onto an iOS mobile device and compare against Google Tango. 

\begin{figure}[t]
    \centering
    \includegraphics[width=0.45\textwidth]{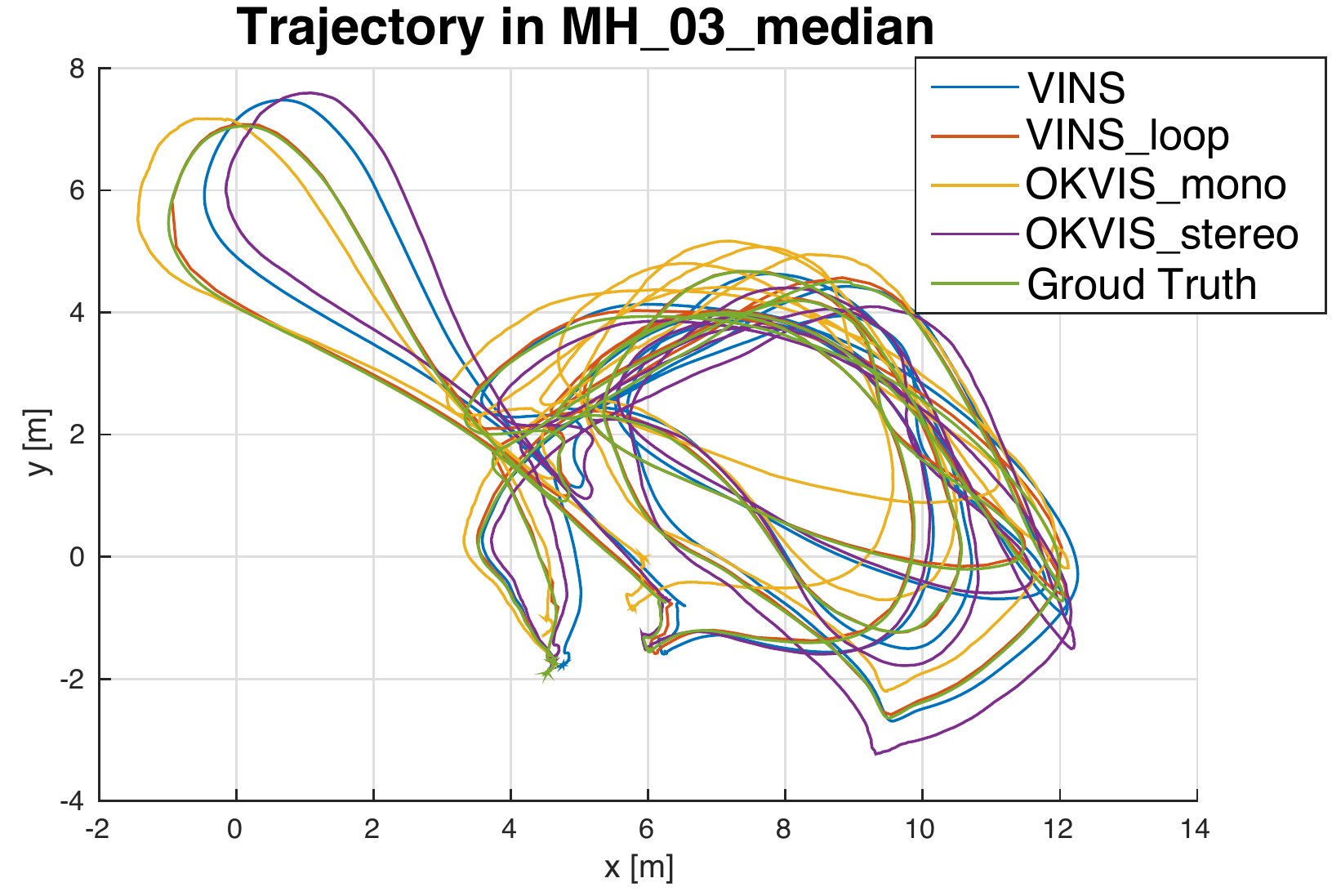}
    \caption{Trajectory in MH\_03\_median, compared with OKVIS.
        \label{fig:MH_03_trajectory}}
\end{figure}

\begin{figure}[h]
    \centering
    \includegraphics[width=0.45\textwidth]{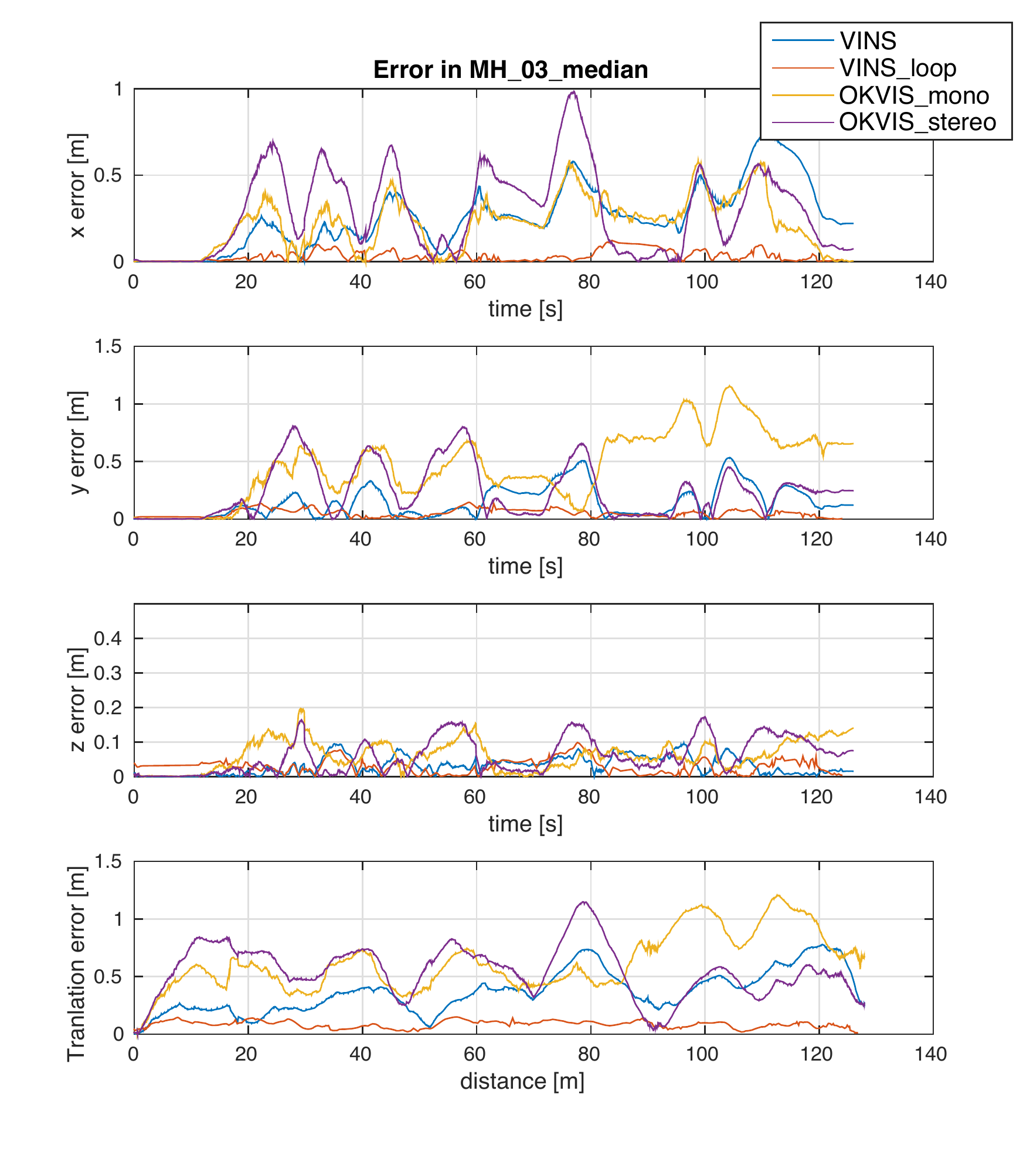}
    \caption{Translation error plot in MH\_03\_median.
        \label{fig:MH_03_error}}
	\vspace{-0.2cm}
\end{figure}

\subsection{Dataset Comparison}

We evaluate our proposed VINS-Mono using the EuRoC MAV Visual-Inertial Datasets~\cite{Burri25012016}. 
The datasets are collected onboard a micro aerial vehicle, 
which contains stereo images (Aptina MT9V034 global shutter, WVGA monochrome, 20 FPS), 
synchronized IMU measurements (ADIS16448, 200 Hz), 
and ground truth states (VICON and Leica MS50). 
We only use images from the left camera. 
Large IMU bias and illumination changes are observed in these datasets. 

In these experiments, we compare VINS-Mono with OKVIS~\cite{LeuFurRab1306}, a state-of-the-art VIO that works with monocular and stereo cameras. 
OKVIS is another optimization-based sliding-window algorithm.
Our algorithm is different with OKVIS in many details, as presented in the technical sections.
Our system is complete with robust initialization and loop closure.
We use two sequences, MH\_03\_median and MH\_05\_difficult, to show the performance of proposed method.
To simplify the notation, we use VINS to denote our approach with only monocular VIO, and VINS\_loop to denote the complete version with relocalization and pose graph optimization. 
We use OKVIS\_mono and OKVIS\_stereo to denote the OKVIS's results using monocular and stereo images respectively.
For the fair comparison, we throw the first 100 outputs, and use the following 150 outputs to align with the ground truth, and compare the remaining estimator outputs.

\begin{figure}[t]
    \centering
    \includegraphics[width=0.42\textwidth]{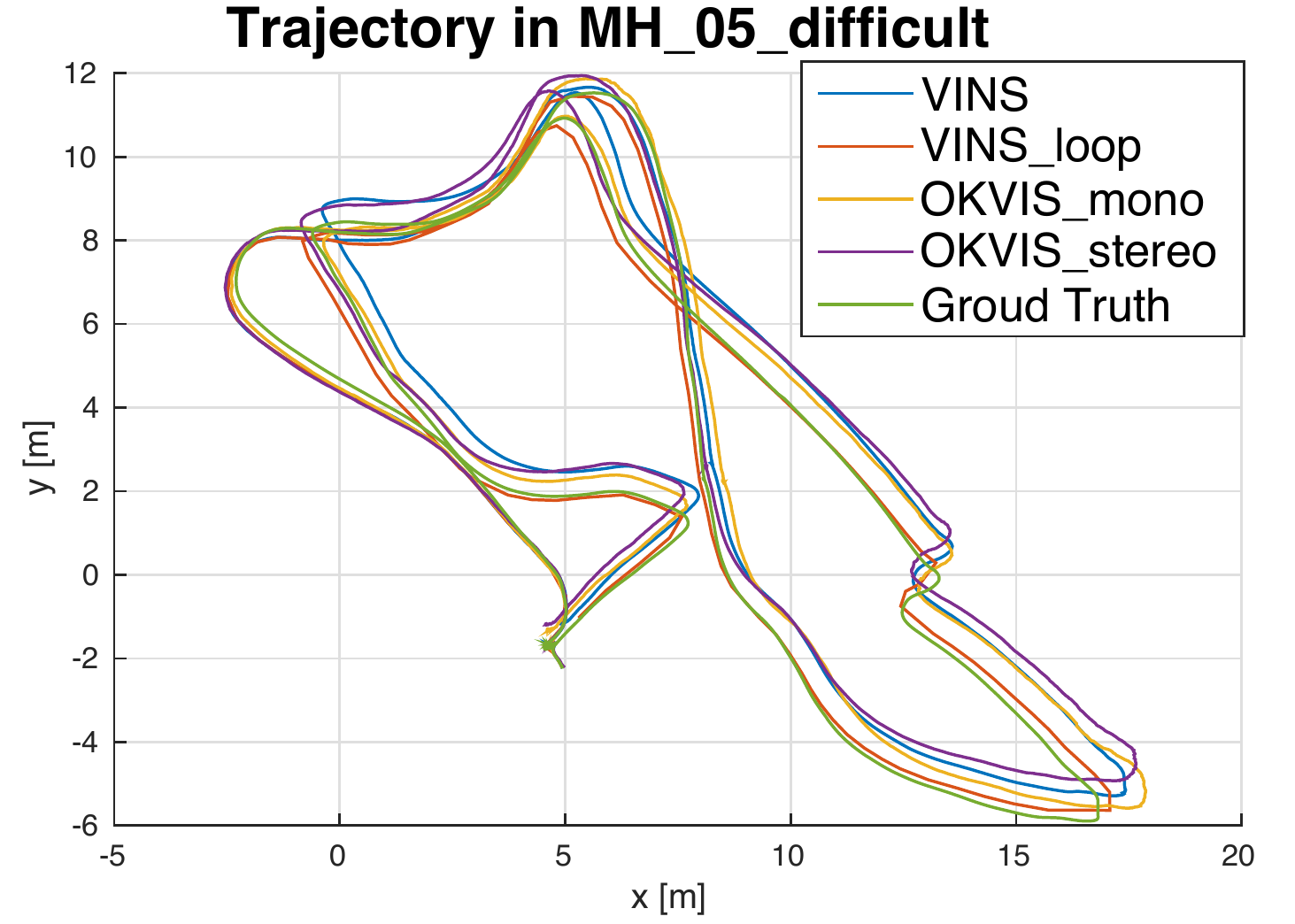}
    \caption{Trajectory in MH\_05\_difficult, compared with OKVIS..
        \label{fig:MH_05_trajectory}}
\end{figure}

\begin{figure}[h]
    \centering
    \includegraphics[width=0.45\textwidth]{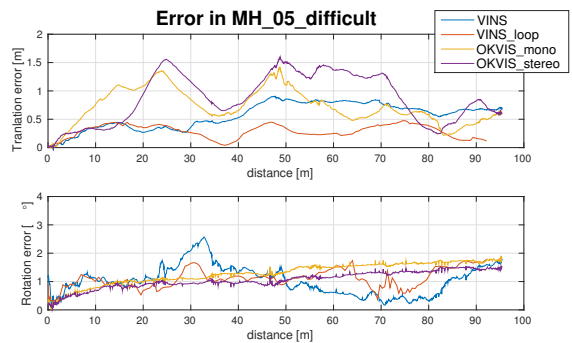}
    \caption{Translation error and rotation error plot in MH\_05\_difficult.
        \label{fig:MH_05_error}}
    \vspace{-0.2cm}
\end{figure}

For the sequence MH\_03\_median, the trajectory is shown in Fig.~\ref{fig:MH_03_trajectory}. 
We only compare translation error since rotated motion is negligible in this sequence.
The x, y, z error versus time, and the translation error versus distance are shown in Fig.~\ref{fig:MH_03_error}. 
In the error plot, VINS-Mono with loop closure has the smallest translation error. 
We observe similar results in MH\_05\_difficult. 
The proposed method with loop function has the smallest translation error.
The translation and rotation errors are shown in Fig.~\ref{fig:MH_05_error}. 
Since the movement is smooth without much yaw angle change in this sequence, only position drift occurs.
Obviously, the loop closure capability efficiently bound the accumulated drifts. 
OKVIS performs better in roll and pitch angle estimation.
A possible reason is that VINS-Mono uses the pre-integration technique which is the first-order approximation of IMU propagation to save computation resource.    

VINS-Mono performs well in all EuRoC datasets, even in the most challenging sequence, V1\_03\_difficult, the one includes aggressive motion, texture-less area, and significant illumination change. 
The proposed method can initialize quickly in V1\_03\_difficult, due to the dedicated initialization procedure.

For pure VIO, both VINS-Mono and OKVIS have similar accuracy, it is hard to distinguish which one is better. 
However, VINS-Mono outperforms OKVIS at the system level. 
It is a complete system with robust initialization and loop closure function to assist the monocular VIO.

\subsection{Indoor Experiment}
\begin{figure}
    \centering
    \includegraphics[width=0.38\textwidth]{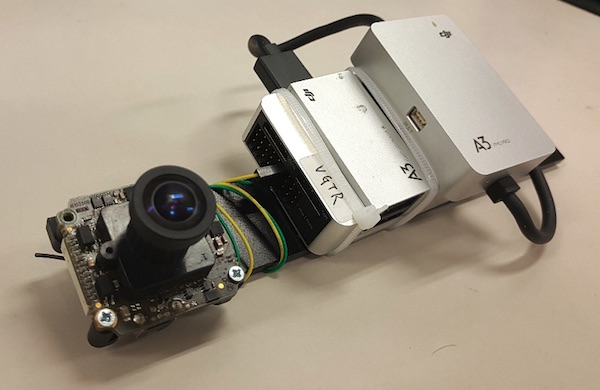}
    \caption{The device we used for the indoor experiment. 
             It contains one forward-looking global shutter camera (MatrixVision mvBlueFOX-MLC200w) with 752$\times$480 resolution.
             We use the built-in IMU (ADXL278 and ADXRS290, 100Hz) for the DJI A3 flight controller.
        \label{fig:device}}
\end{figure}

\begin{figure}
    \centering
    \subfigure[Pedestrians]{
        \label{fig:l1}    
        \includegraphics[width=0.45\columnwidth]{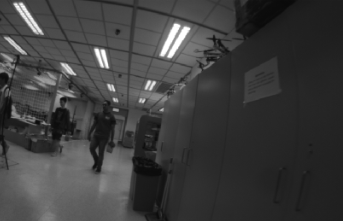}}   
    \subfigure[texture-less area]{
        \label{fig:l2}
        \includegraphics[width=0.45\columnwidth]{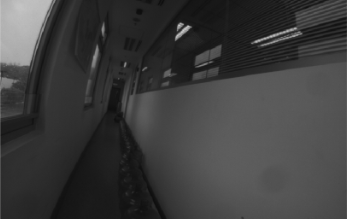}}
    \subfigure[Low light condition]{
        \label{fig:l3}
        \includegraphics[width=0.45\columnwidth]{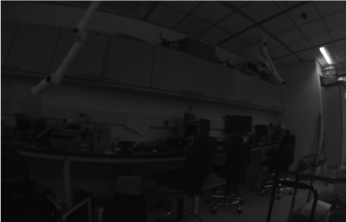}}
    \subfigure[Glass and reflection]{
        \label{fig:l4}
        \includegraphics[width=0.45\columnwidth]{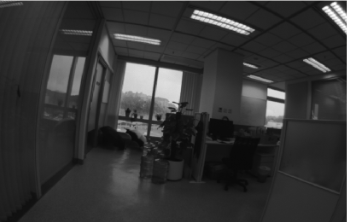}}
    \caption{Sample images for the challenging indoor experiment.}
    \label{fig:loop_environment}
        \vspace{-0.5cm}
\end{figure}

In the indoor experiment, we choose our laboratory environment as the experiment area. 
The sensor suite we use is shown in Fig.~\ref{fig:device}.
It contains a monocular camera (20Hz) and an IMU (100Hz) inside the DJI A3 controller\footnote{\url{http://www.dji.com/a3}}. 
We hold the sensor suite by hand and walk in normal pace in the laboratory. 
We encounter pedestrians, low light condition, texture-less area, glass and reflection, as shown in Fig.~\ref{fig:loop_environment}.
Videos can be found in the multimedia attachment.

\begin{figure}
    \centering
    \subfigure[Trajectory of OKVIS.]{
        \label{fig:indoor_okvis}    
        \includegraphics[width=0.45\columnwidth]{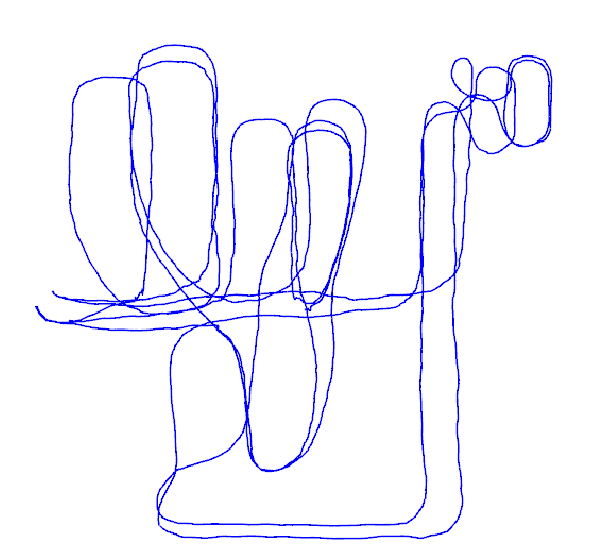}}   
    \subfigure[Trajectory of VINS-Mono without loop closure.]{
        \label{fig:loop/indoor_withoutloop}
        \includegraphics[width=0.45\columnwidth]{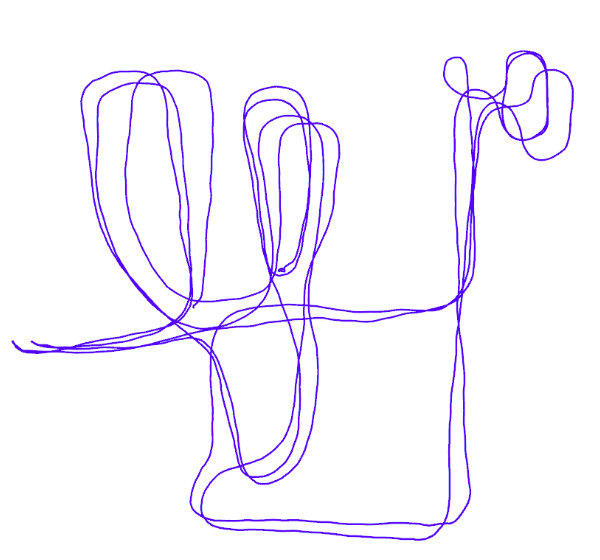}}
        \subfigure[Trajectory of VINS-Mono with relocalization and loop closure. Red lines indicate loop detection.]{
        \label{fig:loop/indoor_loop}
        \includegraphics[width=0.45\columnwidth]{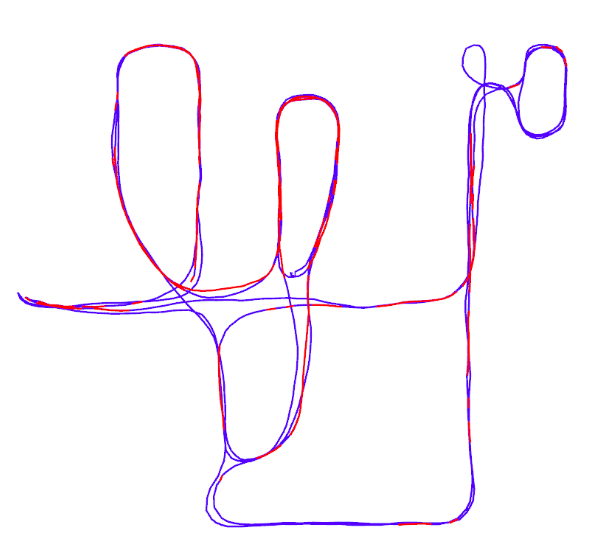}}
    \caption{Results of the indoor experiment with comparison against OKVIS.}
    \label{fig:loop_indoor}
    \vspace{-0.5cm}
\end{figure}

We compare our result with OKVIS, as shown in Fig.~\ref{fig:loop_indoor}. 
Fig.~\ref{fig:indoor_okvis} is the VIO output from OKVIS. 
Fig.~\ref{fig:loop/indoor_withoutloop} is the VIO-only result from proposed method without loop closure. 
Fig.~\ref{fig:loop/indoor_loop} is the result of the proposed method with relocalization and loop closure.
Noticeable VIO drifts occurred when we circle indoor.
Both OKVIS and the VIO-only version of VINS-Mono accumulate significant drifts in x, y, z, and yaw angle. 
Our relocalization and loop closure modules efficiently eliminate all these drifts. 

\subsection{Large-scale Environment}

\subsubsection{Go out of the lab}
We test VINS-Mono in a mixed indoor and outdoor setting. 
The sensor suite is the same as the one shown in Fig.~\ref{fig:device}.
We started from a seat in the laboratory and go around the indoor space. 
Then we went down the stairs and walked around the playground outside the building. 
Next, we went back to the building and go upstairs. Finally, we returned to the same seat in the laboratory.
The whole trajectory is more than 700 meters and last approximately ten minutes.
A video of the experiment can be found in the multimedia attachment.

\begin{figure}[h]
    \centering
    \includegraphics[width=0.4\textwidth]{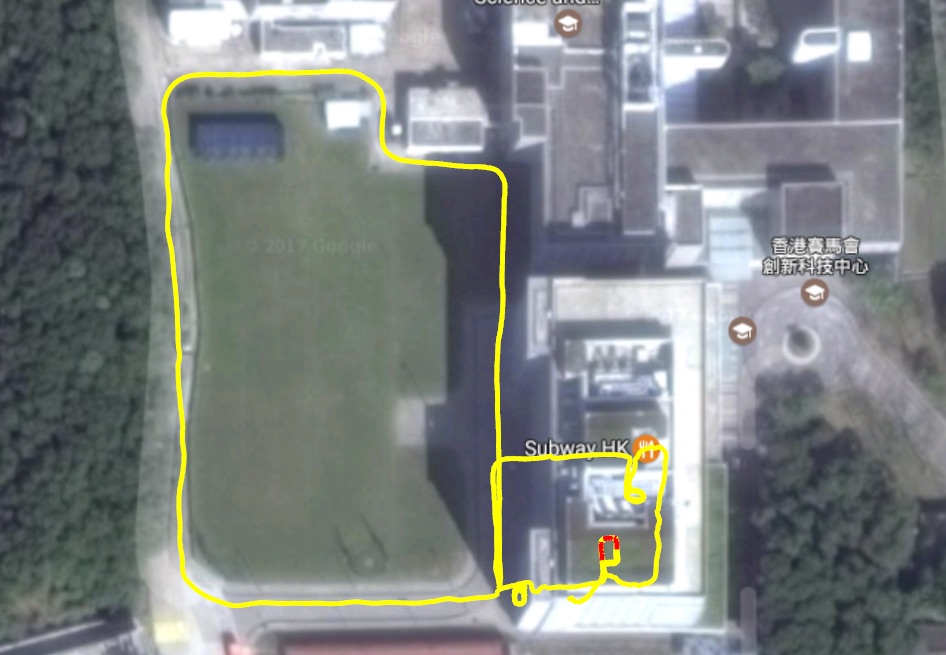}
    \caption{The estimated trajectory of the mixed indoor and outdoor experiment aligned with Google Map. 
             The yellow line is the final estimated trajectory from VINS-Mono after loop closure. 
             Red lines indicate loop closure.
        \label{fig:out_of_lab_map}}
\end{figure}

\begin{figure}
    \centering
        \subfigure[Estimated trajectory from OKVIS]{
            \label{fig:out_of_lab2}    
            \includegraphics[width=0.48\columnwidth]{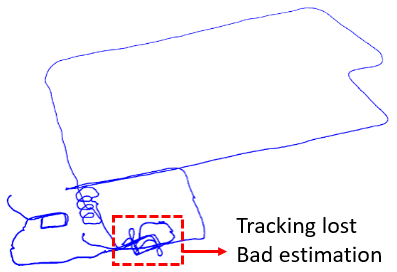}}
    \subfigure[Estimated trajectory from VINS-Mono with loop closure disabled]{
        \label{fig:out_of_lab0}    
        \includegraphics[width=0.45\columnwidth]{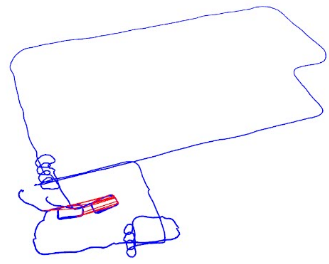}}   
    \subfigure[Estimated trajectory from VINS-Mono with loop closure]{
        \label{fig:out_of_lab1}
        \includegraphics[width=0.45\columnwidth]{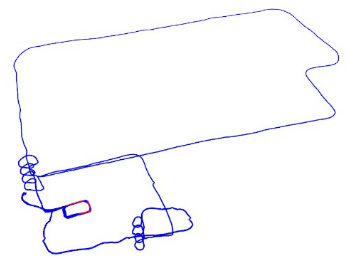}}
    \caption{Estimated trajectories for the mixed indoor and outdoor experiment.
             In Fig.~\ref{fig:out_of_lab2}, results from OKVIS went bad during tracking lost in texture-less region (staircase).
             Figs.~\ref{fig:out_of_lab0}~\ref{fig:out_of_lab1} shows results from VINS-Mono without and with loop closure.
             Red lines indicate loop closures. 
             The spiral-shaped blue line shows the trajectory when going up and down the stairs.
             We can see that VINS-Mono performs well (subject to acceptable drift) even without loop closure.}
    \label{fig:out_of_lab}
\end{figure}

The trajectory is shown in Fig.~\ref{fig:out_of_lab}. 
Fig.~\ref{fig:out_of_lab2} is the trajectory from OKVIS. When we went up stairs, OKVIS shows unstable feature tracking, resulting in bad estimation. We cannot see the shape of stairs in the red block. 
The VIO-only result from VINS-Mono is shown in Fig.~\ref{fig:out_of_lab0}.
The trajectory with loop closure is shown in Fig.~\ref{fig:out_of_lab1}. 
The shape of stairs is clear in proposed method. 
The closed loop trajectory is aligned with Google Map to verify its accuracy, as shown in Fig.~\ref{fig:out_of_lab_map}.  

The final drift for OKVIS is [13.80, -5.26. 7.23]m in x, y and z-axis.
The final dirft of VINS-Mono without loop closure is [-5.47, 2.76. -0.29]m, which occupies 0.88\% with respect to the total trajectory length, smaller than OKVIS 2.36\%.  
With loop correction. the final drift is bounded to [-0.032, 0.09, -0.07]m, which is trivial compared to the total trajectory length. 
Although we do not have ground truth, we can still visually inspect that the optimized trajectory is smooth and can be precisely aligned with the satellite map.

\subsubsection{Go around campus}
\begin{figure*}
    \centering
    \includegraphics[width=0.8\textwidth]{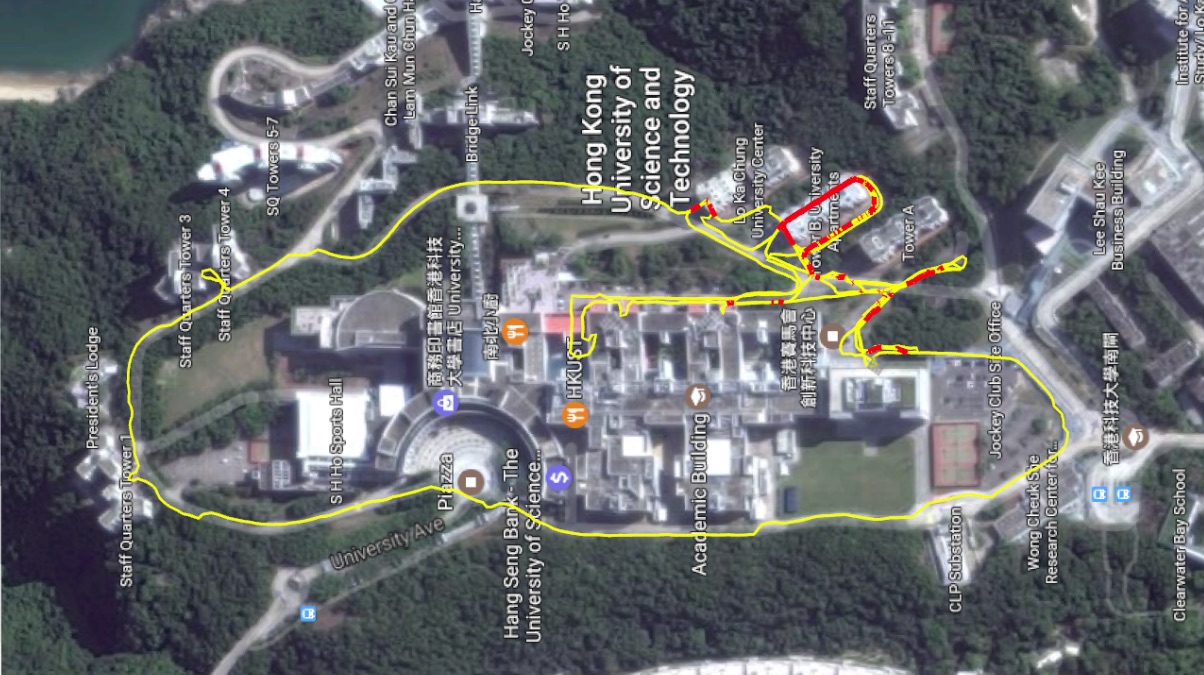}
    \caption{The estimated trajectory of the very large-scale environment aligned with Google map. 
             The yellow line is the estimated trajectory from VINS-Mono.
             Red lines indicates loop closure.
        \label{fig:hkust_map}}
\end{figure*}

This very large-scale dataset that goes around the whole HKUST campus was recorded with a handheld VI-Sensor\footnote{http://www.skybotix.com/}.
The dataset covers the ground that is around 710m in length, 240m in width, and with 60m in height changes. 
The total path length is 5.62km.
The data contains the 25Hz image and 200Hz IMU lasting for 1 hour and 34 minutes.
It is a very significant experiment to test the stability and durability of VINS-Mono.

In this large-scale test, We set the keyframe database size to 2000 in order to provide sufficient loop information and achieve real-time performance.
We run this dataset with an Intel i7-4790 CPU running at 3.60GHz.  
Timing statistics are show in Table.~\ref{table:time}.
The estimated trajectory is aligned with Google map in Fig.~\ref{fig:hkust_map}.
Compared with Google map, we can see our results are almost drift-free in this very long-duration test.

\begin{table}[b]
    \caption{\label{table:time} Timing Statistics}
    \begin{tabular}{cccc}
        \toprule
        \textbf{Tread}& \textbf{Modules}            & \textbf{Time (ms)} & \textbf{Rate (Hz)}   \\
        \midrule
        \multirow{2}{*}{1} & Feature detector & 15 & 25 \\
        & KLT tracker & 5 & 25 \\
        \midrule
        2 & Window optimization  & 50 & 10 \\
        \midrule
        \multirow{2}{*}{3} & Loop detection & 100 &  \\
        & Pose graph optimization & 130 &  \\
        \bottomrule
    \end{tabular}
\end{table}

\subsection{Application I: Feedback Control of an Aerial Robot}
\label{sec:onboard flight}

\begin{figure}
    \centering
    \subfigure[Aerial robot testbed]{
        \label{fig:px}    
        \includegraphics[width=0.38\textwidth]{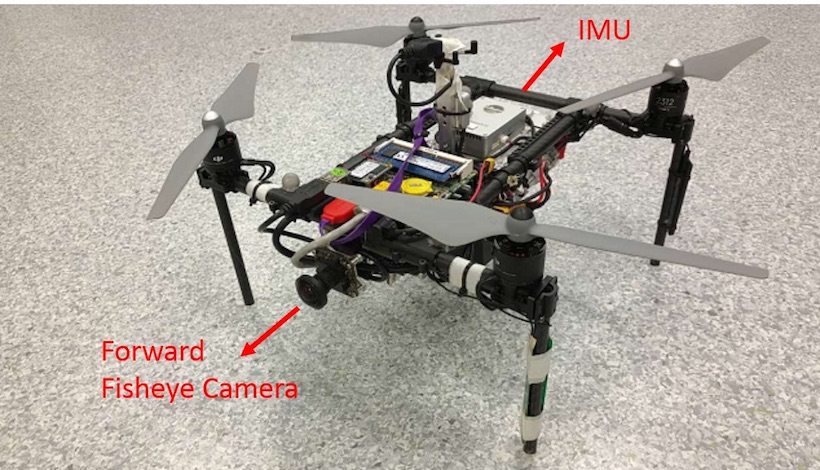}}   
    \subfigure[Testing environment and desired figure eight pattern]{
        \label{fig:px_indoor}
        \includegraphics[width=0.38\textwidth]{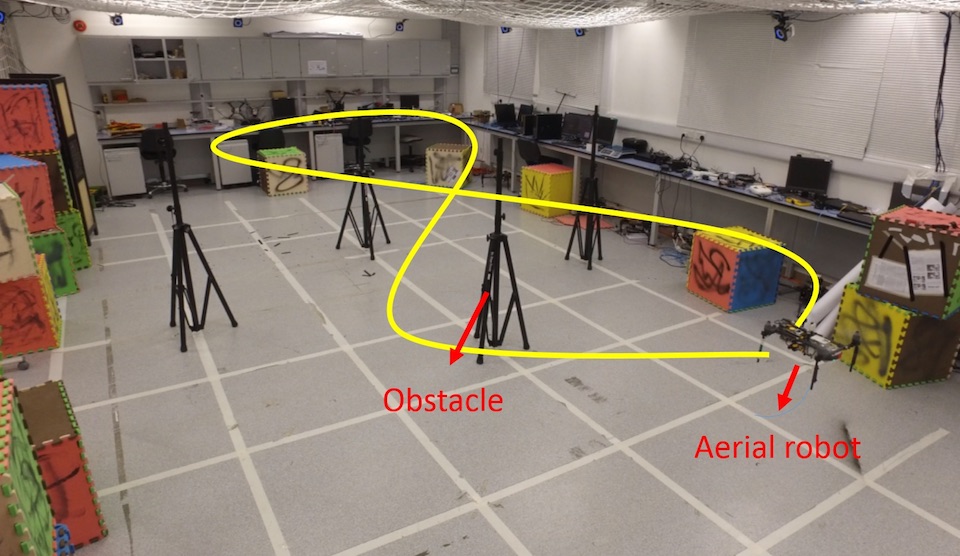}}
    \caption{Fig~\ref{fig:px}: The self-developed aerial robot with a forward-looking fisheye camera (MatrixVision mvBlueFOX-MLC200w, 190 FOV) and an DJI A3 flight controller (ADXL278 and ADXRS290, 100Hz). 
             Fig.~\ref{fig:px_indoor}: The designed trajectory. Four known obstacles are placed. 
             The yellow line is the predefined figure eight-figure pattern which the aerial robot should follow. 
             The robot follows the trajectory four times with loop closure disabled. 
             A video of the experiment can be found in the multimedia attachment.}
    \label{fig:px_experiment}
    \vspace{-0.5cm}
\end{figure}

We apply VINS-Mono for autonomous feedback control of an aerial robot, as shown in Fig.~\ref{fig:px}. 
We use a forward-looking global shutter camera (MatrixVision mvBlueFOX-MLC200w) with 752$\times$480 resolution, and equipped it with a 190-degree fisheye lens.
A DJI A3 flight controller is used for both IMU measurements and for attitude stabilization control.
The onboard computation resource is an Intel i7-5500U CPU running at 3.00 GHz.
Traditional pinhole camera model is not suitable for large FOV camera.  
We use MEI~\cite{mei2007single} model for this camera, calibrated by the toolkit introduced in~\cite{heng2013camodocal}.

In this experiment, we test the performance of autonomous trajectory tracking under using state estimates from VINS-Mono. 
Loop closure is disabled for this experiment
The quadrotor is commanded to track a figure eight pattern with each circle being 1.0 meters in radius, as shown in Fig.~\ref{fig:px_indoor}. 
Four obstacles are put around the trajectory to verify the accuracy of VINS-Mono without loop closure. 
The quadrotor follows this trajectory four times continuously during the experiment. 
The 100 Hz onboard state estimates (Sect.~\ref{subsec:imu_propagation}) enables real-time feedback control of quadrotor.

\begin{figure*}
    \centering
    \includegraphics[width=0.93\textwidth]{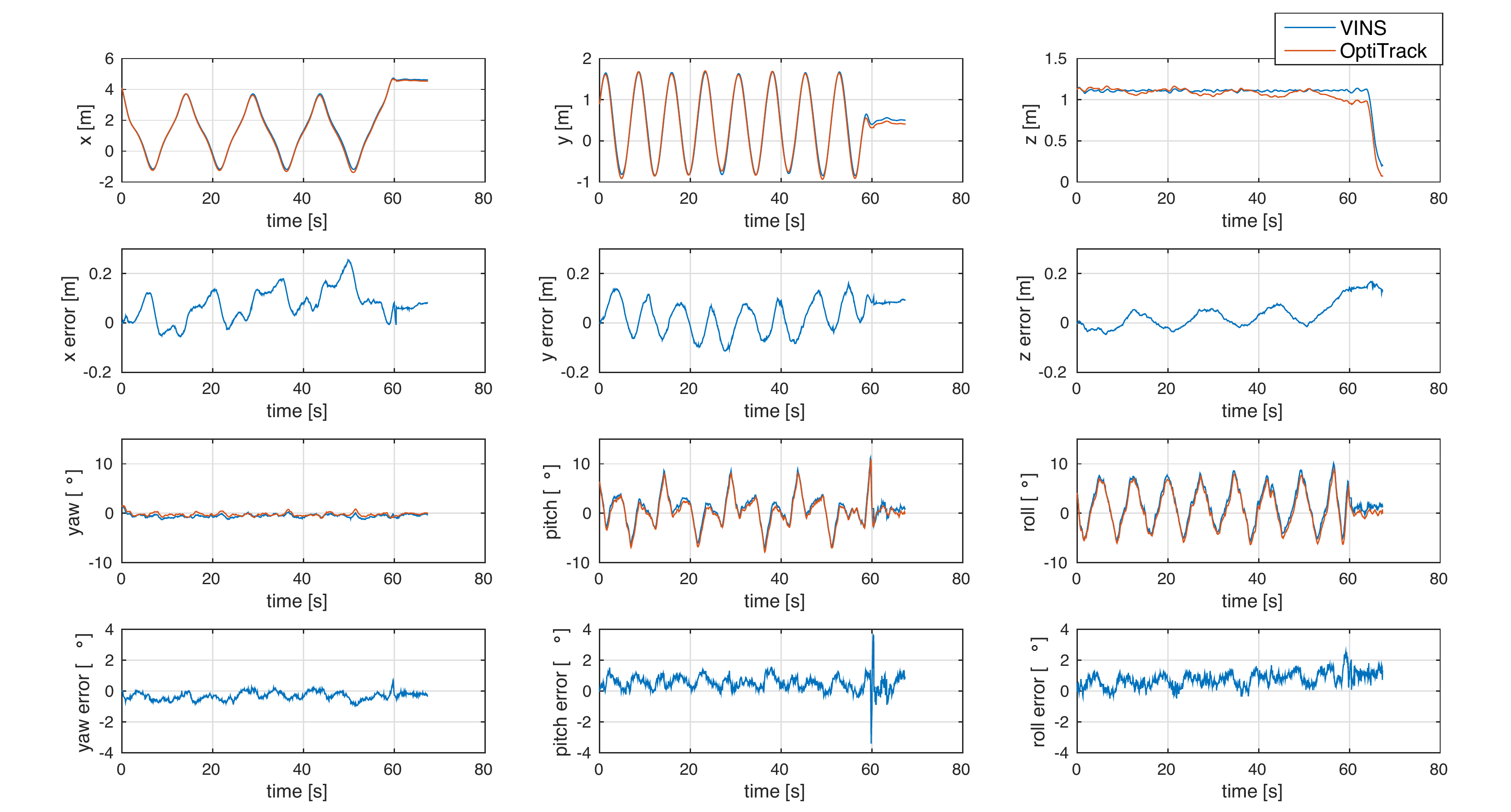}
    \caption{Position, orientation and their corresponding errors of loop closure-disabled VINS-Mono compared with OptiTrack. 
             A sudden in pitch error at the 60s is caused by aggressive breaking at the end of the designed trajectory, and possible time misalignment error between VINS-Mono and OptiTrack.
        \label{fig:eight_details}}
\end{figure*}

Ground truth is obtained using OptiTrack\footnote{http://optitrack.com/}.
Total trajectory length is 61.97 m.
The final drift is [0.08, 0.09, 0.13] m, resulting in 0.29\% position drift.
Details of the translation and rotation as well as their corresponding errors are shown in Fig.~\ref{fig:eight_details}. 

\begin{figure}
    \centering
    \includegraphics[width=0.45\textwidth]{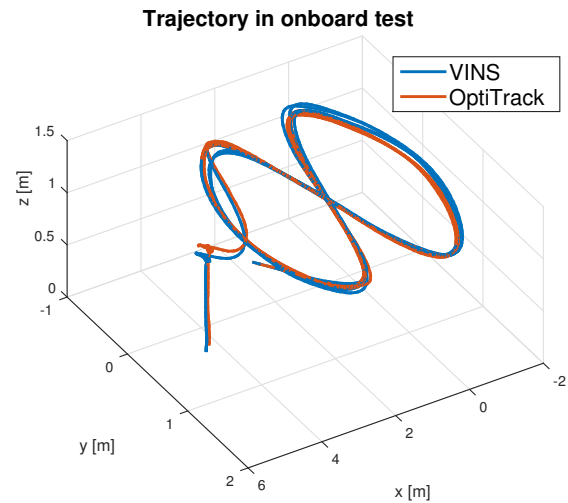}
    \caption{The trajectory of loop closure-disabled VINS-Mono on the MAV platform and its comparison against the ground truth. 
             The robot follows the trajectory four times. 
             VINS-Mono estimates are used as real-time position feedback for the controller. 
             Ground truth is obtained using OptiTrack. Total length is $61.97m$. Final drift is $0.18m$.
        \label{fig:vins_trajectory}}
        \vspace{-0.5cm}
\end{figure}

\subsection{Application II: Mobile Device}

We port VINS-Mono to mobile devices and present a simple AR application to showcase its accuracy and robustness. 
We name our mobile implementation VINS-Mobile\footnote{https://github.com/HKUST-Aerial-Robotics/VINS-Mobile}, and we compare it against Google Tango device\footnote{http://shopap.lenovo.com/hk/en/tango/}, 
which is one of the best commercially available augmented reality solutions on mobile platforms.

\begin{figure}
    \centering   
    \includegraphics[width=0.7\columnwidth]{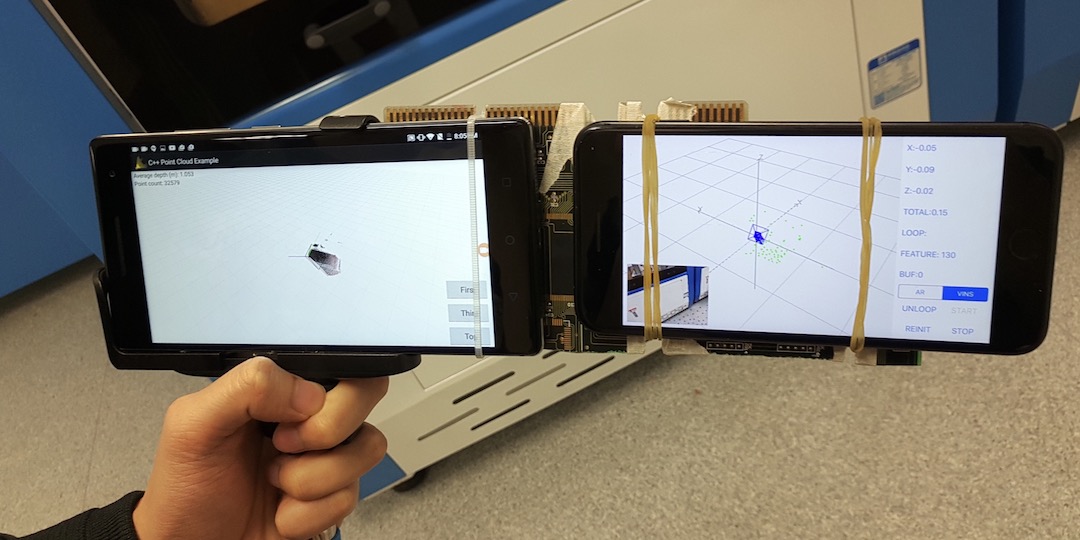}
    \caption{A simple holder that we used to mount the Google Tango device (left) and the iPhone7 Plus (right) that runs our VINS-Mobile implementation.}
    \label{fig:device_mobile} 
\end{figure}

\begin{figure}[h]
    \centering 
    \subfigure[Beginning] { \label{fig:compare1} 
        \includegraphics[width=0.9\columnwidth]{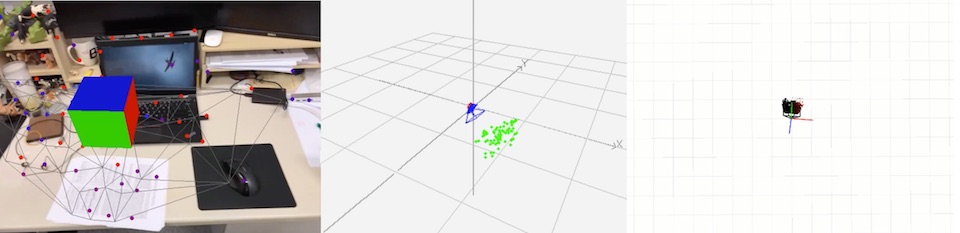} 
    } 
    \subfigure[Door opening] { \label{fig:compare2} 
        \includegraphics[width=0.9\columnwidth]{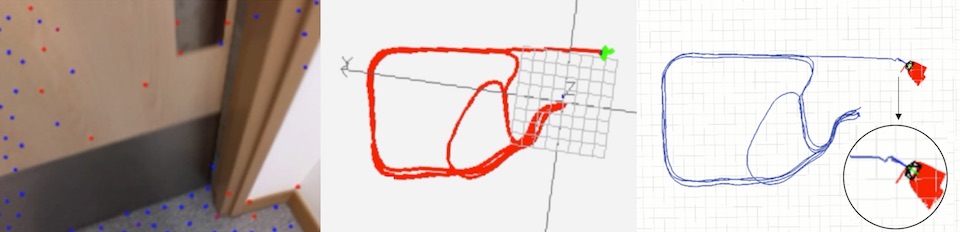} 
    } 
    \subfigure[End] { \label{fig:compare3} 
        \includegraphics[width=0.9\columnwidth]{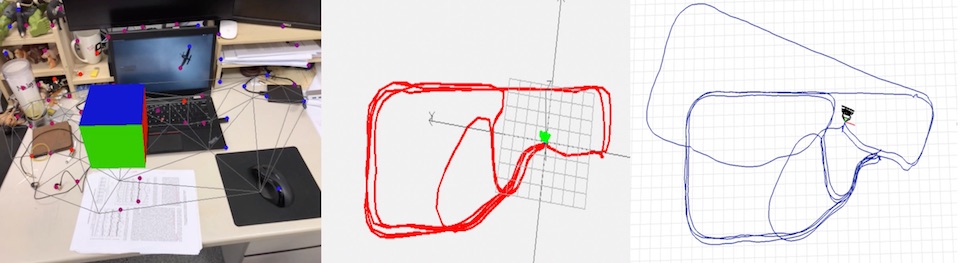} 
    }
    \caption{From left to right: AR image from VINS-Mobile, estimated trajectory from VINS-Mono, estimated trajectory from Tango 
    Fig:~\ref{fig:compare1}: Both VINS-Mobile and Tango are initialized at the start location and a virtual box is inserted on the plane which extracted from estimated features. 
    Fig.~\ref{fig:compare2}: A challenging case in which the camera is facing a moving door. The drift of Tango trajectory is highlighted. 
    Fig.~\ref{fig:compare3}: The final trajectory of both VINS-Mobile and Tango. The total length is about 264m.}
    \label{fig:compare} 
    \vspace{-0.2cm}
\end{figure}

VINS-Mono runs on an iPhone7 Plus.
we use 30 Hz images with 640 $\times$ 480 resolution captured by the iPhone, and IMU data at 100 Hz obtained by the built-in InvenSense MP67B 6-axis gyroscope and accelerometer.  
As Fig.~\ref{fig:device_mobile} shows, we mount the iPhone together with a Tango-Enabled Lenovo Phab 2 Pro.
The Tango device uses a global shutter fisheye camera and synchronized IMU for state estimation.
Firstly, we insert a virtual cube on the plane which is extracted from estimated visual features as shown in Fig.~\ref{fig:compare1}. 
Then we hold these two devices and walk inside and outside the room in a normal pace. 
When loops are detected, we use the 4-DOF pose graph optimization (Sect.~\ref{sec:pose_graph}) to eliminate x, y, z and yaw drifts.

Interestingly, when we open a door, Tango's yaw estimation jumps a big angle, as shown in Fig.~\ref{fig:compare2}. 
The reason maybe estimator crash caused by unstable feature tracking or active failure detection and recovery. 
However, VINS-Mono still works well in this challenging case. 
After traveling about 264m, we return to the start location. 
The final result can be seen in Fig.~\ref{fig:compare3}, the trajectory of Tango suffers drifting in the last lap while our VINS returns to the start point.
The drift in total trajectory is eliminated due to the 4-DOF pose graph optimization.
This is also evidenced by the fact that the cube is registered to the same place on the image comparing to the beginning. 

Admittedly, Tango is more accurate than our implementation especially for local state estimates.
However, this experiment shows that the proposed method can run on general mobile devices and have the potential ability to compare specially engineered devices.
The robustness of proposed method is also proved in this experiment.
Video can be found in the multimedia attachment.

\section{Conclusion and Future Work}
\label{sec:conclusion}
In this paper, we propose a robust and versatile monocular visual-inertial estimator.
Our approach features both state-of-the-art and novel solutions to IMU pre-integration, estimator initialization and failure recovery, online extrinsic calibration, tightly-coupled visual-inertial odometry, relocalization, and efficient global optimization.
We show superior performance by comparing against both state-of-the-art open source implementations and highly optimized industry solutions.
We open source both PC and iOS implementation for the benefit of the community.

Although feature-based VINS estimators have already reached the maturity of real-world deployment, we still see many directions for future research.
Monocular VINS may reach weakly observable or even degenerate conditions depending on the motion and the environment. 
We are most interested in online methods to evaluate the observability properties of monocular VINS, and online generation of motion plans to restore observability.
Another research direction concerns the mass deployment of monocular VINS on a large variety of consumer devices, such as mobile phones.
This application requires online calibration of almost all sensor intrinsic and extrinsic parameters, as well as the online identification of calibration qualities.
Finally, we are interested in producing dense maps given results from monocular VINS. Our first results on monocular visual-inertial dense mapping with application to drone navigation was presented in~\cite{LinGaoQinGaoLiuWuYangShen17}.
However, extensive research is still necessary to further improve the system accuracy and robustness.


\ifCLASSOPTIONcaptionsoff
  \newpage
\fi

\bibliographystyle{IEEEtran}
\bibliography{paper}

\begin{thebibliography}{10}
\providecommand{\url}[1]{#1}
\csname url@samestyle\endcsname
\providecommand{\newblock}{\relax}
\providecommand{\bibinfo}[2]{#2}
\providecommand{\BIBentrySTDinterwordspacing}{\spaceskip=0pt\relax}
\providecommand{\BIBentryALTinterwordstretchfactor}{4}
\providecommand{\BIBentryALTinterwordspacing}{\spaceskip=\fontdimen2\font plus
\BIBentryALTinterwordstretchfactor\fontdimen3\font minus
  \fontdimen4\font\relax}
\providecommand{\BIBforeignlanguage}[2]{{%
\expandafter\ifx\csname l@#1\endcsname\relax
\typeout{** WARNING: IEEEtran.bst: No hyphenation pattern has been}%
\typeout{** loaded for the language `#1'. Using the pattern for}%
\typeout{** the default language instead.}%
\else
\language=\csname l@#1\endcsname
\fi
#2}}
\providecommand{\BIBdecl}{\relax}
\BIBdecl

\bibitem{klein2007parallel}
G.~Klein and D.~Murray, ``Parallel tracking and mapping for small ar
  workspaces,'' in \emph{Mixed and Augmented Reality, 2007. ISMAR 2007. 6th
  IEEE and ACM International Symposium on}.\hskip 1em plus 0.5em minus
  0.4em\relax IEEE, 2007, pp. 225--234.

\bibitem{ForPizSca1405}
C.~Forster, M.~Pizzoli, and D.~Scaramuzza, ``{SVO}: Fast semi-direct monocular
  visual odometry,'' in \emph{Proc. of the {IEEE} Int. Conf. on Robot. and
  Autom.}, Hong Kong, China, May 2014.

\bibitem{engel2014lsd}
J.~Engel, T.~Sch{\"o}ps, and D.~Cremers, ``Lsd-slam: Large-scale direct
  monocular slam,'' in \emph{European Conference on Computer Vision}.\hskip 1em
  plus 0.5em minus 0.4em\relax Springer International Publishing, 2014, pp.
  834--849.

\bibitem{mur2015orb}
R.~Mur-Artal, J.~Montiel, and J.~D. Tardos, ``Orb-slam: a versatile and
  accurate monocular slam system,'' \emph{IEEE Transactions on Robotics},
  vol.~31, no.~5, pp. 1147--1163, 2015.

\bibitem{engel2017direct}
J.~Engel, V.~Koltun, and D.~Cremers, ``Direct sparse odometry,'' \emph{IEEE
  Transactions on Pattern Analysis and Machine Intelligence}, 2017.

\bibitem{GalvezTRO12}
D.~G\'alvez-L\'opez and J.~D. Tard\'os, ``Bags of binary words for fast place
  recognition in image sequences,'' \emph{IEEE Transactions on Robotics},
  vol.~28, no.~5, pp. 1188--1197, October 2012.

\bibitem{SheMicKum1505}
S.~Shen, N.~Michael, and V.~Kumar, ``Tightly-coupled monocular visual-inertial
  fusion for autonomous flight of rotorcraft {MAV}s,'' in \emph{Proc. of the
  {IEEE} Int. Conf. on Robot. and Autom.}, Seattle, WA, May 2015.

\bibitem{yang2017monocular}
Z.~Yang and S.~Shen, ``Monocular visual--inertial state estimation with online
  initialization and camera--imu extrinsic calibration,'' \emph{IEEE
  Transactions on Automation Science and Engineering}, vol.~14, no.~1, pp.
  39--51, 2017.

\bibitem{QinShen17}
T.~Qin and S.~Shen, ``Robust initialization of monocular visual-inertial
  estimation on aerial robots.'' in \emph{Proc. of the {IEEE/RSJ} Int. Conf. on
  Intell. Robots and Syst.}, Vancouver, Canada, 2017, accepted.

\bibitem{LiShen17}
P.~Li, T.~Qin, B.~Hu, F.~Zhu, and S.~Shen, ``Monocular visual-inertial state
  estimation for mobile augmented reality.'' in \emph{Proc. of the {IEEE} Int.
  Sym. on Mixed adn Augmented Reality}, Nantes, France, 2017, accepted.

\bibitem{weiss2012real}
S.~Weiss, M.~W. Achtelik, S.~Lynen, M.~Chli, and R.~Siegwart, ``Real-time
  onboard visual-inertial state estimation and self-calibration of mavs in
  unknown environments,'' in \emph{Robotics and Automation (ICRA), 2012 IEEE
  International Conference on}.\hskip 1em plus 0.5em minus 0.4em\relax IEEE,
  2012, pp. 957--964.

\bibitem{lynen2013robust}
S.~Lynen, M.~W. Achtelik, S.~Weiss, M.~Chli, and R.~Siegwart, ``A robust and
  modular multi-sensor fusion approach applied to mav navigation,'' in
  \emph{Proc. of the {IEEE/RSJ} Int. Conf. on Intell. Robots and Syst.}\hskip
  1em plus 0.5em minus 0.4em\relax IEEE, 2013, pp. 3923--3929.

\bibitem{MouRou0704}
A.~I. Mourikis and S.~I. Roumeliotis, ``A multi-state constraint {K}alman
  filter for vision-aided inertial navigation,'' in \emph{Proc. of the {IEEE}
  Int. Conf. on Robot. and Autom.}, Roma, Italy, Apr. 2007, pp. 3565--3572.

\bibitem{LiMou1305}
M.~Li and A.~Mourikis, ``High-precision, consistent {EKF}-based visual-inertial
  odometry,'' \emph{Int. J. Robot. Research}, vol.~32, no.~6, pp. 690--711, May
  2013.

\bibitem{bloesch2015robust}
M.~Bloesch, S.~Omari, M.~Hutter, and R.~Siegwart, ``Robust visual inertial
  odometry using a direct ekf-based approach,'' in \emph{Proc. of the
  {IEEE/RSJ} Int. Conf. on Intell. Robots and Syst.}\hskip 1em plus 0.5em minus
  0.4em\relax IEEE, 2015, pp. 298--304.

\bibitem{LeuFurRab1306}
S.~Leutenegger, S.~Lynen, M.~Bosse, R.~Siegwart, and P.~Furgale,
  ``Keyframe-based visual-inertial odometry using nonlinear optimization,''
  \emph{Int. J. Robot. Research}, vol.~34, no.~3, pp. 314--334, Mar. 2014.

\bibitem{mur2016visual}
R.~Mur-Artal and J.~D. Tardos, ``Visual-inertial monocular {SLAM} with map
  reuse,'' \emph{arXiv preprint arXiv:1610.05949}, 2016.

\bibitem{wu2015square}
K.~Wu, A.~Ahmed, G.~A. Georgiou, and S.~I. Roumeliotis, ``A square root inverse
  filter for efficient vision-aided inertial navigation on mobile devices.'' in
  \emph{Robotics: Science and Systems}, 2015.

\bibitem{paulcomparative}
M.~K. Paul, K.~Wu, J.~A. Hesch, E.~D. Nerurkar, and S.~I. Roumeliotis, ``A
  comparative analysis of tightly-coupled monocular, binocular, and stereo
  vins,'' in \emph{Proc. of the {IEEE} Int. Conf. on Robot. and Autom.},
  Singapore, May 2017.

\bibitem{kaess2012isam2}
M.~Kaess, H.~Johannsson, R.~Roberts, V.~Ila, J.~J. Leonard, and F.~Dellaert,
  ``isam2: Incremental smoothing and mapping using the bayes tree,'' \emph{The
  International Journal of Robotics Research}, vol.~31, no.~2, pp. 216--235,
  2012.

\bibitem{usenko2016direct}
V.~Usenko, J.~Engel, J.~St{\"u}ckler, and D.~Cremers, ``Direct visual-inertial
  odometry with stereo cameras,'' in \emph{Proc. of the {IEEE} Int. Conf. on
  Robot. and Autom.}\hskip 1em plus 0.5em minus 0.4em\relax IEEE, 2016, pp.
  1885--1892.

\bibitem{LupSuk1202}
T.~Lupton and S.~Sukkarieh, ``Visual-inertial-aided navigation for high-dynamic
  motion in built environments without initial conditions,'' \emph{{IEEE}
  Trans. Robot.}, vol.~28, no.~1, pp. 61--76, Feb. 2012.

\bibitem{ForCarDel1507}
C.~Forster, L.~Carlone, F.~Dellaert, and D.~Scaramuzza, ``{IMU} preintegration
  on manifold for efficient visual-inertial maximum-a-posteriori estimation,''
  in \emph{Proc. of Robot.: Sci. and Syst.}, Rome, Italy, Jul. 2015.

\bibitem{SheMulMic1406}
S.~Shen, Y.~Mulgaonkar, N.~Michael, and V.~Kumar, ``Initialization-free
  monocular visual-inertial estimation with application to autonomous {MAVs},''
  in \emph{Proc. of the Int. Sym. on Exp. Robot.}, Marrakech, Morocco, Jun.
  2014.

\bibitem{Mar1308}
A.~Martinelli, ``Closed-form solution of visual-inertial structure from
  motion,'' \emph{Int. J. Comput. Vis.}, vol. 106, no.~2, pp. 138--152, 2014.

\bibitem{kaiser2017simultaneous}
J.~Kaiser, A.~Martinelli, F.~Fontana, and D.~Scaramuzza, ``Simultaneous state
  initialization and gyroscope bias calibration in visual inertial aided
  navigation,'' \emph{IEEE Robotics and Automation Letters}, vol.~2, no.~1, pp.
  18--25, 2017.

\bibitem{faessler2015automatic}
M.~Faessler, F.~Fontana, C.~Forster, and D.~Scaramuzza, ``Automatic
  re-initialization and failure recovery for aggressive flight with a monocular
  vision-based quadrotor,'' in \emph{Proc. of the {IEEE} Int. Conf. on Robot.
  and Autom.}\hskip 1em plus 0.5em minus 0.4em\relax IEEE, 2015, pp.
  1722--1729.

\bibitem{strasdat2010scale}
H.~Strasdat, J.~Montiel, and A.~J. Davison, ``Scale drift-aware large scale
  monocular slam,'' \emph{Robotics: Science and Systems VI}, 2010.

\bibitem{LucKan8108}
B.~D. Lucas and T.~Kanade, ``An iterative image registration technique with an
  application to stereo vision,'' in \emph{Proc. of the Intl. Joint Conf. on
  Artificial Intelligence}, Vancouver, Canada, Aug. 1981, pp. 24--28.

\bibitem{shi1994good}
J.~Shi and C.~Tomasi, ``Good features to track,'' in \emph{Computer Vision and
  Pattern Recognition, 1994. Proceedings CVPR'94., 1994 IEEE Computer Society
  Conference on}.\hskip 1em plus 0.5em minus 0.4em\relax IEEE, 1994, pp.
  593--600.

\bibitem{hartley2003multiple}
R.~Hartley and A.~Zisserman, \emph{Multiple view geometry in computer
  vision}.\hskip 1em plus 0.5em minus 0.4em\relax Cambridge university press,
  2003.

\bibitem{heyden2005multiple}
A.~Heyden and M.~Pollefeys, ``Multiple view geometry,'' \emph{Emerging Topics
  in Computer Vision}, 2005.

\bibitem{nister2004efficient}
D.~Nist{\'e}r, ``An efficient solution to the five-point relative pose
  problem,'' \emph{IEEE transactions on pattern analysis and machine
  intelligence}, vol.~26, no.~6, pp. 756--770, 2004.

\bibitem{LiuShen17}
T.~Liu and S.~Shen, ``Spline-based initialization of monocular visual-inertial
  state estimators at high altitude,'' \emph{{IEEE} Robotics and Automation
  Letters}, 2017, accepted.

\bibitem{lepetit2009epnp}
V.~Lepetit, F.~Moreno-Noguer, and P.~Fua, ``Epnp: An accurate o (n) solution to
  the pnp problem,'' \emph{International journal of computer vision}, vol.~81,
  no.~2, pp. 155--166, 2009.

\bibitem{triggs1999bundle}
B.~Triggs, P.~F. McLauchlan, R.~I. Hartley, and A.~W. Fitzgibbon, ``Bundle
  adjustment—a modern synthesis,'' in \emph{International workshop on vision
  algorithms}.\hskip 1em plus 0.5em minus 0.4em\relax Springer, 1999, pp.
  298--372.

\bibitem{Hub64}
P.~Huber, ``Robust estimation of a location parameter,'' \emph{Annals of
  Mathematical Statistics}, vol.~35, no.~2, pp. 73--101, 1964.

\bibitem{ceres-solver}
S.~Agarwal, K.~Mierle, and Others, ``Ceres solver,''
  \url{http://ceres-solver.org}.

\bibitem{SibMatSuk1009}
G.~Sibley, L.~Matthies, and G.~Sukhatme, ``Sliding window filter with
  application to planetary landing,'' \emph{J. Field Robot.}, vol.~27, no.~5,
  pp. 587--608, Sep. 2010.

\bibitem{calonder2010brief}
M.~Calonder, V.~Lepetit, C.~Strecha, and P.~Fua, ``Brief: Binary robust
  independent elementary features,'' \emph{Computer Vision--ECCV 2010}, pp.
  778--792, 2010.

\bibitem{Burri25012016}
M.~Burri, J.~Nikolic, P.~Gohl, T.~Schneider, J.~Rehder, S.~Omari, M.~W.
  Achtelik, and R.~Siegwart, ``The euroc micro aerial vehicle datasets,''
  \emph{The International Journal of Robotics Research}, 2016.

\bibitem{mei2007single}
C.~Mei and P.~Rives, ``Single view point omnidirectional camera calibration
  from planar grids,'' in \emph{Robotics and Automation, 2007 IEEE
  International Conference on}.\hskip 1em plus 0.5em minus 0.4em\relax IEEE,
  2007, pp. 3945--3950.

\bibitem{heng2013camodocal}
L.~Heng, B.~Li, and M.~Pollefeys, ``Camodocal: Automatic intrinsic and
  extrinsic calibration of a rig with multiple generic cameras and odometry,''
  in \emph{Intelligent Robots and Systems (IROS), 2013 IEEE/RSJ International
  Conference on}.\hskip 1em plus 0.5em minus 0.4em\relax IEEE, 2013, pp.
  1793--1800.

\bibitem{LinGaoQinGaoLiuWuYangShen17}
Y.~Lin, F.~Gao, T.~Qin, W.~Gao, T.~Liu, W.~Wu, Z.~Yang, and S.~Shen,
  ``Autonomous aerial navigation using monocular visual-inertial fusion,''
  \emph{J. Field Robot.}, vol.~00, pp. 1--29, 2017.

\end{thebibliography}

\end{document}